\documentclass[compsoc,twocolumn,journal]{mytran}
\RequirePackage[compress]{cite}
\RequirePackage{bbold}
\RequirePackage{verbatim}
\RequirePackage{colortbl}
\RequirePackage{graphicx}
\RequirePackage{subfig}
\RequirePackage{verbatim}
\RequirePackage{amssymb}
\RequirePackage{amsmath}
\RequirePackage{amsthm}
\RequirePackage{multirow}
\RequirePackage{rotating}
\RequirePackage{wrapfig}
\RequirePackage{placeins}
\RequirePackage{afterpage}
\definecolor{menucolor}{rgb}{0.1,0.52,0.47}
\definecolor{anchorcolor}{rgb}{0.85,0.37,0.01}
\definecolor{runcolor}{rgb}{0.46,0.44,0.701}
\definecolor{linkcolor}{rgb}{0.3,0.55,0.01}
\definecolor{urlcolor}{rgb}{0.12,0.47,0.70}
\definecolor{citecolor}{rgb}{0.55,0.36,0.01}
\definecolor{filecolor}{rgb}{0.4,0.4,0.4}
\makeatletter
\newcommand{\pushright}[1]{\ifmeasuring@#1\else\omit\hfill$\displaystyle#1$\fi\ignorespaces}
\newcommand{\pushleft}[1]{\ifmeasuring@#1\else\omit$\displaystyle#1$\hfill\fi\ignorespaces}
\makeatother

\newcommand{\mAR}{\mathcal AR}
\newcommand{\mD}{\mathcal D}

\newcommand{\bmm}{\boldsymbol{m}}

\newcommand{\bx}{\boldsymbol{x}}
\newcommand{\bX}{\boldsymbol{X}}
\newcommand{\by}{\boldsymbol{y}}
\newcommand{\bY}{\boldsymbol{Y}}

\newcommand{\mJ}{\mathcal J}

\newcommand{\mR}{\mathcal R}

\newcommand{\ben}{\begin{enumerate}}
\newcommand{\een}{\end{enumerate}}

\usepackage{multirow}
\usepackage{booktabs}

\usepackage{amsmath, amsfonts, mathtools, amsthm,physics,amssymb}

\usepackage{subfig}
\usepackage{adjustbox}

\newcommand{\citep}{\cite}
\newcommand{\citet}{\cite}

\usepackage{color,array,verbatim,algpseudocode,bm}
\definecolor{menucolor}{rgb}{0.1,0.52,0.47}
\definecolor{urlcolor}{rgb}{0.85,0.37,0.01}
\definecolor{runcolor}{rgb}{0.46,0.44,0.701}
\definecolor{filecolor}{rgb}{0.2,0.5,0.01}
\definecolor{linkcolor}{rgb}{0.12,0.47,0.70}
\definecolor{citecolor}{rgb}{0.55,0.36,0.01}
\definecolor{anchorcolor}{rgb}{0.4,0.4,0.4}
\usepackage[colorlinks=true, urlcolor=urlcolor, linkcolor=linkcolor,citecolor=citecolor,filecolor=filecolor, anchorcolor=anchorcolor,menucolor=menucolor]{hyperref}
\usepackage{nameref}
\usepackage{zref-xr}
\zxrsetup{
  tozreflabel=false,
  toltxlabel=true,
}
\makeatletter
\newcommand\VeryLarge{\@setfontsize\Huge{16}{16}}
\newcommand\smaller{\@setfontsize\small{4}{4}}
\makeatother   

\def\subparagraph{} % because IEEE classes don't define this, but titlesec assumes it's present
\usepackage{titlesec}
\titlespacing*{\section}{0pt}{*1}{*1}
\titlespacing{\subsection}{0pt}{*1}{*1}
\titlespacing{\subsubsection}{0pt}{*1}{*1}

\titleformat{\paragraph}[runin]{\normalfont\bfseries}{\theparagraph}{0.5em}{}[:]

\IEEEaftertitletext{\vspace{-3\baselineskip}} 



\renewcommand{\baselinestretch}{0.88}
\normalsize

\begin{document}
\title{
  \vspace{-0.2in}
  \VeryLarge CatSIM: A Categorical Image Similarity Metric}
\author{
  \vspace{-0.15in}
  Geoffrey Z. Thompson and Ranjan Maitra\\
  Department of Statistics, Iowa State University, Ames, Iowa 50011, USA.
\\ \vspace{-0.4in}
%\thanks{{\bf Abbreviations:} WSS, within-sums-of-squares; $k_k$-means: kernel $k$-means; OT, optimal transfer; QT, quick transfer; AAC, alternating assignment and centering; OTQT, Optimal transfer and quick transfer; SRSWOR, simple random sampling without replacement;  MKNN, mean $k$-nearest neighbors; MATR, MAx-TRace; SDSS, Sloan Digital Sky Survey; ARI, adjusted Rand Index; VIF, visual information fidelity; SCAC, sampling-based clustering and classification.}
}
%\vspace{-0.7in}
\maketitle
%\IEEEcompsoctitleabstractindextext{
  \vspace{0.1in}
\begin{abstract}
We introduce CatSIM, a new similarity metric for binary and   multinary two- and three-dimensional images and volumes. CatSIM uses a structural similarity image quality
  paradigm and is robust to small perturbations in location so that structures in similar, but not entirely overlapping, image or volumetric regions are rated higher than by simple matching. The metric can also compare arbitrary regions inside images and volumes. CatSIM is evaluated  on artificial data sets, validated by comparing with human perception in two separate image quality  assessment surveys, and illustrated  %real classification and segmentation data. 
  on
two datasets. % and in the assessment of the
%  agreement of classification systems.
  The publicly available R package  \texttt{catsim} implements the methodology.
\end{abstract}
\vspace{-0.08in}
%}
%\section{Introduction}

\section{Introduction}
Similarity metrics are important for comparing
images and image volumes, for example, when evaluating 
image processing or segmentation algorithms. Categorical images assign
nominal values at each pixel/voxel, such as in tissue or land
classification in medical or satellite images.
%discrete colors, classification in satellite images,  tissue status
%in medical images, etc.
Most categorical image similarity methods only
provide pointwise class comparisons at each
pixel/voxel~\citep{Baulieu1989,Gower1986} and ignore spatial context. Such
measures~\citep{taha2014} include   
the Jaccard ($\mJ$)~\citep{jaccard1901} or Dice ($\mD$)~\citep{Dice1945} indices
for binary problems, Cohen's 
$\kappa$~\citep{cohen1960} or Hamming's distance~\citep{hamming50} for
multi-class comparisons when labels have the same meaning
in both cases being compared, or the Rand ($\mR$)~\citep{Rand1971} or Adjusted Rand
indices ($\mAR$)~\citep{hubertandarabie85} for when they do
not. These metrics %ignore spatial structure 
%. These include the Jaccard ($\mJ$)~\cite{jaccard1901,jaccard1912} or
%Dice indices~\citep{Dice1945,sorensen1948} for 2-class comparisons,
%or the Adjusted Rand index ($\mAR$)~\cite{Rand1971} for multi-class
%images  considerstatistics and data science in imaging only point-wise agreement or
%disagreement~\cite{Baulieu1989,Gower1986} and ignore the spatial
%structure of the image data. However, and
can be misleading, especially when 
%which is a major deficiency, because in analyzing the images, 
the features of interest are objects or fine structures like lines
that may be spatially perturbed with little visual effect
but large pointwise %pixel-to-pixel or voxel-to-voxel
perturbations. %comparisons to disagree strongly. 

There exist image similarity metrics that  %have been developed to
account for the relationships of nearby points or features. %to each other
%in their assessment. %One example application of this is evaluating
                   %different image segmentations and comparing them
                   %to a ground truth segmentation~\cite{taha2014}. 
For instance, \citet{huttenlocher1993} and \citet{Prieto2003} use geometric methods %for binary and grayscale images
that account for spatial and intensity distortions by identifying and
comparing object edges in images while \citet{russakoffetal04}
uses region-wise calculations and asymptotic
normality arguments to arrive at a regional mutual information metric
for comparing two images. %by considering regions rather than points~\cite{russakoffetal04}.
%Perhaps the most popular comparison metric is the %The highly-popular
The popular multiscale structural similarity (SSIM) or
MS-SSIM~\citep{wang2003multiscale, 
  wang2004image} for color and grayscale images accounts for
spatial and intensity distortions as well as structural 
information across multiple scales in the image. A computationally
intensive version, CW-SSIM~\citep{sampatwang09} %, has been developed for grayscale and binary images %that is invariant to  small geometric distortions 
exists for grayscale and binary images.  These methods 
and adaptations of them to different domains such as medical imaging generally have
good performance but do not always align with human
assessment~\citep{Renieblas2017, Mason2020}. \begin{comment}
Instead of simple statistics, CW-SSIM puts the
magnitude of the complex wavelets of an image through a SSIM-like
algorithm before multiplication by a penalizing constant to evaluate
how how far out of phase it is from the phase image. The CW-SSIM also
doesn't have to perform the
downsampling  stages because that's
implicit in the construction of the
wavelets (as a "6-scale
16-directional orientable
pyramid"). It is also much more
computationally complex.
\end{comment}
Such methods also do not apply to multi-class images or volumes so we 
propose (Section~\ref{sec:meth}) methodology that  adapts statistics
specifically appropriate for multinary and binary data to an
SSIM-like approach.  % and that looks at  regions of an image at
% multiple scales in order to capture similarities in the structure of the image in a way that is generalizable to three-dimensional maps.
Section~\ref{sec:illustrate} illustrates and validates our
methods. The paper concludes with some discussion. Additional details are in appendices. %sections, figures and tables referenced here using  the prefix ``S'' is available. The methodology is implemented in 
The R package {\tt catsim}, publicly available on the Comprehensive R Archive Network (CRAN), implements our methodology.% http://dx.doi.org/10.1109/TMI.2019.xxxxxx

\section{Methodology}\label{sec:meth}
% needed in second column of first page if using \IEEEpubid
%\IEEEpubidadjcol
\subsection{ Background and Previous Work} %Structural Similarity Indices}
%Structural Similarity Index~
SSIM~\citep{wang2003multiscale}
is an image quality assessment index conceptually designed to account 
for structural similarities in images as visualized by a
human rater. %aims to mimic assessments by the human visual system. 
The basic version of the index combines summary statistics  on sliding
$N\!\!\times\!\!N$ aligned patches of the images ($\bX$ and $\bY$) being
compared. % and combinesthem together for an overall index. 
Let $\bx$ and $\by$ be aligned patches from $\bX$ and $\bY$, with 
$\mu_{\bx}$ and $\mu_{\by}$ being the averaged values in each patch, $\sigma_{\bx}^2$ and
$\sigma_{\by}^2$ the variances, and $\sigma_{\bx\by}$ the covariance.
The SSIM is calculated from the luminance, contrast and structural similarity
functions, $l(\bx,\by)$, $c(\bx,\by)$ and $s(\bx,\by)$ defined as follows. The function 
$l(\bx,\by) = f(\mu_{\bx},\mu_{\by};C_1)$ and $c(\bx,\by) =
f(\sigma_{\bx},\sigma_{\by};C_2)$  where
\begin{equation*}\label{eqn:ssim}
  f(\theta,\phi;k) = \frac{2\theta\phi+k}{\theta^2+\phi^2+k},
\end{equation*}
and 
\begin{equation*}
s(\bx,\by)
                                                                =
                                                                \frac{\sigma_{\bx\by}+C_3}{\sigma_{\bx}\sigma_{\by}+C_3},
\end{equation*}
with $C_1, C_2,$ and $C_3$ small constants. The three  
functions take values in [0,1], with 1 attained for identical
patches. The means and the standard deviations (SDs) of the two
patches influence $l(\bx,\by)$  %luminance and contrast functions  
and $c(\bx,\by)$ while $s(\bx,\by)$ %the structural
% similarity function
compares the covariance to their individual SDs. Averaging each function
over all possible $N\!\times\!N$ patches in $\bX$ and $\bY$ provides
$l(\bX,\bY)$, $c(\bX,\bY)$ and $s(\bX,\bY)$. Multiplying these quantities yields
$\textrm{SSIM}(\bX,\bY) = l(\bX,\bY)  c(\bX,\bY)  s(\bX, \bY)$.

The MS-SSIM~\citep{wang2003multiscale} enhancement computes SSIM at multiple scales after
downsampling and combines the results from each scale. Specifically,
\begin{equation*} %\label{eqn:prod}
  \textrm{MS-SSIM}(\bX,\bY) {=} l_M(\bX,\bY)^{\alpha_M} \prod_{j = 1}^M c_j(\bX,\bY)^{\beta_j}s_j(\bX,\bY)^{\gamma_j}
  \end{equation*}
  where $j$ indexes the scale at which $c(\bX,\bY)$ and $s(\bX,\bY)$
  are computed, % the corresponding statistic was computed on,
  $M$ is the highest scale after $M{-}1$ iterations of re-scaling, and
  $\alpha_j$, 
  $\beta_j$, and $\gamma_j$  are constants on the $M$-dimensional
  simplex. \citet{wang2003multiscale} uses $\alpha_j{=}\beta_j{=}\gamma_j$
  with empirically specified values. The metric aims to capture 
  local variation and structural similarities between images on
  several scales in a way that mimics the human eye.
  
The CW-SSIM~\citep{sampatwang09} metric uses the product of functions of the
  magnitude and phase of the  complex wavelet coefficients of images
  downsampled over 6 levels. The magnitudes  feed into a SSIM index that  is unity when both images are the same. The metric uses a function of the dot product of the phases  that is 0 only when both images are aligned. Further, CW-SSIM is said to 
  tie to computational models for vision~\citep{sampatwang09} and has the underlying premise that local wavelet   coefficients have consistent phase changes with certain image   distortions but  do not change the structural content of the image. 

  The SSIM and MS-SSIM methods can be made to work for binary images
  (by considering 0-1 to be part of the continuum) while CW-SSIM is
  formulated for binary or grayscale images. However, neither extends
  directly to nominal multi-class images. One could somewhat crudely use one-hot encoding of each category, and apply either of these methods to each individual category, but how the results should be combined is not completey clear, especially with adaptive  downsampling~\citep{marin19}.
  CW-SSIM can  also not
  handle image data with missing observations (such as pixels/voxels  outside
  a mask) as noted in~\citet{malpica2009} in their development of R-SSIM for range images. Also, while not a major limitation, neither of the methods are currently implemented for 3D  volumes. So we develop a SSIM metric for multi-class images and volumes. 
\subsection{Development of CatSIM}

The SSIM philosophy can be developed for multinary images by
defining suitable binary or multinary analogs of the stages of
the MS-SSIM algorithm. In Section~\ref{subsec:index}, we
introduce luminance, contrast, and structural
similarity functions that respect the multinary nature of the data.
In Section~\ref{subsec:downsample}, we define the method of
downsampling to other scales. Finally, in Section~\ref{subsec:algorithm},
we specify how the results are combined across scales
to produce the final metric. %Here, the MS-SSIM approach of using a  low-pass filter
               %and then downsampling by a factor of two is
               %inapplicable because it disregards the structure of
               %the data in the binary case and is meaningless for
               %multi-class image volumes.
%We address each of these challenges next. 

\subsubsection{Index Functions}\label{subsec:index}
We first define statistics for a categorical image or image patch
$\bx {=} (x_1, x_2, \ldots, x_n)$ and  $x_i {\in} \{1, 2, \ldots, K\}$.
Let $\bmm_{\bx}$ be the vector of the proportions of each class in the
patch $\bx$, 
% $\bmm_Y$ the same for $y$, and
and $S_{\bx}$ be a categorical variance measure~\citep{Allaj2017}. That
is, define
\begin{align*}
    p_i &= \frac1n \# \{x_j = i,\, j \in 1, 2, \dots, n \} \quad \mathrm{for\, } i = 1, 2, \ldots, K\\
  \bmm_{\bx} &= \{p_i\}_{i=1}^K,\quad
  S_{\bx} = \frac{1-\sqrt{\sum_{i=1}^K p_i^2 }}{1-1/\sqrt{K}} = \frac{1-\|\bmm_{\bx}\|_2}{1-1/\sqrt{K}}.
\end{align*}
These quantities are related to the Gini-Simpson
index~\citep{Simpson1949, Ceriani2011} but are
scaled similarly to the SD %rather than the variance
and constrained to be in $[0,1]$.
For two aligned $N_u{\times}N_v ({\times} N_w)$ sections of $K$-class
image (volume) patches $\bx$ and $\by$, with $\bmm_{\bx}$,
$\bmm_{\by}$, $S_{\bx}$, and $S_{\by}$, we construct luminance,
contrast, and structural similarity functions analogous to SSIM
for the continuous case. Specifically, we insert categorical analogues
in place of the means and covariances (in SSIM) to get:
\begin{align*}
l^c(\bx,\by) &= \frac{2\bmm_{\bx}^\top \bmm_{\by}+ C_1}{\bmm_{\bx}^\top \bmm_{\bx}  + \bmm_{\by}^\top\bmm_{\by}  + C_1}\\
  c^c(\bx,\by) &= \frac{2S_{\bx}S_{\by} + C_2}{S_{\bx}^2 + S_{\by}^2 + C_2},\quad \mbox{and} \quad s^c(\bx,\by) = v(\bx,\by) ,
\end{align*}
where $C_1$, $C_2,$ are small scalar constants chosen for numerical
stability when the denominator approaches zero, $C_1$ can safely be set to
$0$ and $c^c$ can be redefined to equal to $1$ when the denominator is $0$ to avoid
these constants. % A similar constant
                                %may be inserted into $v(\bx,\by)$ if
                                %needed.
and $v(\bx,\by)$ is an inter-rater agreement measure chosen based on 
the characteristics of the image. For instance, the
$\mJ$  and $\mD$ indices are
appropriate choices for $v(\bx,\by)$ in binary problems where 
presence is more important than absence~\citep{LEVANDOWSKY1971}, while %Cohen's kappa~\cite{cohen1960}, or the Adjusted Rand index~\cite{hubert1985}.
accuracy or $\kappa$~\citep{cohen1960} is more appropriate if
labels have  meaning  and $\mR$ or $\mAR$~\citep{Rand1971,hubertandarabie85} is more appropriate for images with
nominal labels. 
%If $p_o$ is the proportion of $\bx$ that agree with $\by$, in terms
%of the quantities defined here, $\kappa = 1 - (1-p_o)/(1-\bmm_X \cdot
%\bmm_Y)$.  Note that
(Both  $\kappa$ and $\mAR$ can take negative values, and so they are truncated to be in $[0,1]$.)
Further, $C_1$, $C_2,$ can be chosen to be 0 if the indices involved do not allow for a vanishing denominator. %Our implementation of our measure uses $\kappa$ by default.
%be less than 0, so their range should be truncated to remain in
%$[0,1]$. The (unadjusted) Rand index can also be used, as can the
%percentage of pixels agreeing, though neither of these account in any
%way for the probability of chance agreement between two
%classifications. The Dice index, which can be expressed as a function
%of the Jaccard index, can also be used on binary problems.
Our index can accommodate missing values, as in the case of imaging a
volume with known boundaries ({\em e.g.}, a mask in medical imaging
applications), as in Sections~\ref{subsec:phantom} or \ref{sec:fmri}, by the
pair-wise deletion of corresponding points in the sliding window
calculations at each level.

\subsubsection{Downsampling and combining across multiple scales}\label{subsec:downsample}
There are two issues to address here, specifically how to perform down-sampling in a
non-continuous (nominal discrete) setting, and also how to combine the different results across scales. The MS-SSIM
algorithm downsamples by a factor of two after using a low-pass
filter to reduce aliasing 
artifacts but that method can not be applied here because it disregards the
structure of the data in the binary case and is meaningless for
multinary images. 
%\cite{decenciere2011} provide down-sampling methods that preserve the
%topological and other characteristics of binary data but do not have
%a clear extension to multinary data.
We propose using the mode of each
$2 {\times} 2 ({\times} 2)$ slice (block) of pixels (with ties broken randomly for multiple modes). The multiple scales can be
combined for categorical images in a similar manner as MS-SSIM. We now detail the specific steps in our algorithm.
 
\subsection{The CatSIM Algorithm}\label{subsec:algorithm}
By default, we specify  uniform window sizes  $N_u{=}N_v{=} 11$ for 2D images and
$N_u=N_v=N_w {=} 5$ for 3D volumes. We also set $M{=}5$ levels and, as in~\citet{wang2003multiscale},
$\alpha_j{=}\beta_j{=}\gamma_j$ $\forall j$. We choose $\gamma_j$s to be
uniform over the $M$-dimensional simplex. These parameters can all be set
based on the application.
\begin{enumerate}
\item\label{step1} For two images $\bX$ and $\bY$, the $c^c(\bx,\by)$ and
  $s^c(\bx,\by)$ statistics 
    are computed over a rolling $N_u{\times}N_v({\times}N_w)$ pixel (voxel)  window and averaged
for the entire image while $l^c(\bx,\by)$ is computed for the base
level. 
\item\label{step2} Downsample each image by a factor of $2$ using the
  mode (breaking ties at random) of each   $2{\times} 2 ({\times} 2)$ block.
\item\label{step3} Repeat Steps~\ref{step1} and~\ref{step2} for each 
  of $M$ total   levels. 
\item Let $l_1^c(\bX,\bY)$,  $c_j^c(\bX,\bY)$ and  $s_j^c(\bX,\bY)$ 
  be the average of $l_1^c(\bx,\by)$, $c_j^c(\bx,\by)$ and  $s_j^c(\bx,\by)$
    over all   $N_u{\times}N_v({\times} N_w)$ blocks, for
  $j=1,2,\ldots,M$. Define 
  \begin{equation}
    \begin{split}
      &\textrm{CatSIM}(\bX,\bY) \\
      & = [l_{1}^c(\bX,\bY)]^{\alpha_M}
      \prod_{j = 1}^M
      c_{j}^c(\bX,\bY)^{\beta_j}s_j^c(\bX,\bY)^{\gamma_j},
    \end{split}
    \end{equation}\label{eq:catsim}
  with $j$ indexing the level at which the luminance, contrast and
    similarity functions are  calculated. Analogous to MS-SSIM, the luminance is only     considered on the top level because it measures the proportions in each class which largely remain unchanged between layers. % and
                                % the $\alpha, \beta,$ and $\gamma$   are defined above.
\end{enumerate}
%For three-dimensional applications, the 3-D image can either be treated as a stack of 2-D images with each slice treated separately and their final results averaged or as a true isotropic 3-D image with all three dimensions treated equally.

%%% Local Variables:
%%% mode: latex
%%% TeX-master: "catsim-paper"
%%% End:
\section{Performance Evaluations}
\label{sec:illustrate}
\subsection{Illustration of CatSIM}\label{subsec:illustrate}
We illustrate CatSIM's behavior on different distortions
and degradations of two binary images (one with  a mask and with highly
unequal class sizes) and one multinary image.
\begin{comment}
Spatial displacement which preserves shapes
results in a high CatSIM index while added salt-and-pepper noise to match the error rate
yields a low CatSIM index, while other metrics rate both highly. The weighting of the
different downsampled layers controls how much this difference is penalized.
\end{comment}
\subsubsection{An expanded Besag binary image}\label{subsec:illustrationbin}
\begin{figure*}[h]
  \centering
\mbox{
  \hspace{-0.015\textwidth}
\subfloat[True]{\includegraphics[width=.141\textwidth]{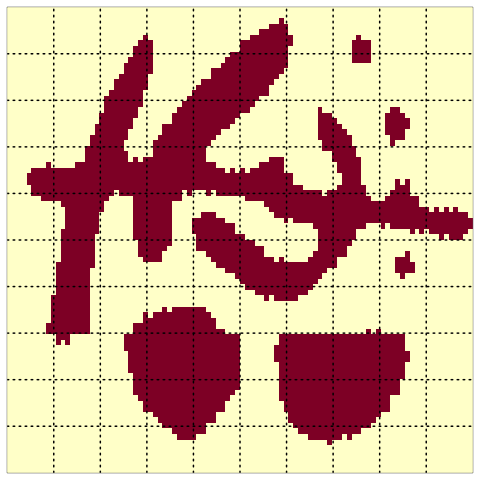}
\label{fig:binsubfig1}}%
%\end{minipage}%
%\qquad
%\begin{minipage}{0.45\textwidth}%
\hspace{-0.005\textwidth}
\subfloat[H.-shift]{\includegraphics[width=0.141\textwidth]{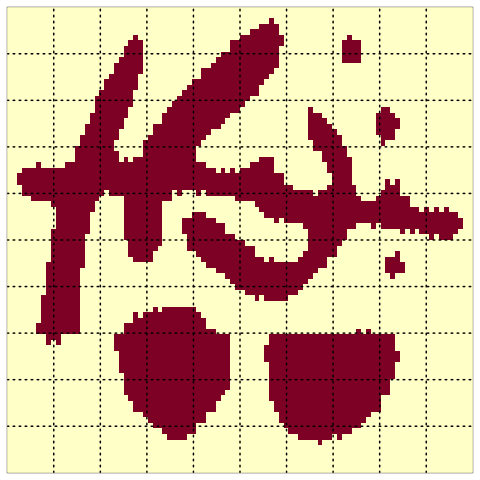}
\label{fig:binsubfig2}}%
\hspace{-0.01\textwidth}
\subfloat[H.-S\&P]{
\includegraphics[width=0.141\textwidth]{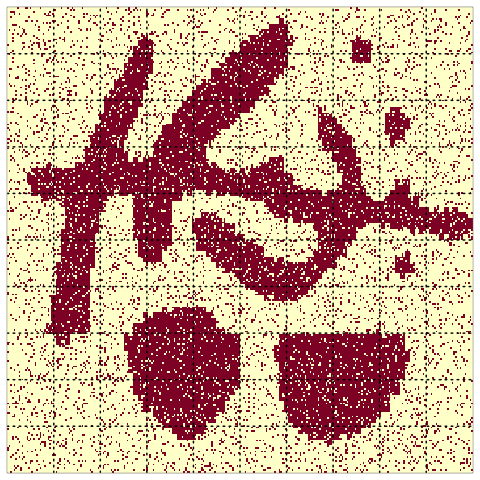}
\label{fig:binsubfig5}}%
\hspace{-0.01\textwidth}
\subfloat[V.-shift]{
\includegraphics[width=0.141\textwidth]{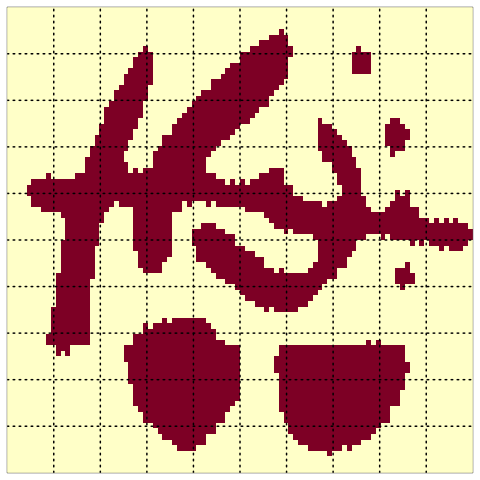}
\label{fig:binsubfig3}}%
\hspace{-0.01\textwidth}
\subfloat[V.-S\&P]{
\includegraphics[width=0.141\textwidth]{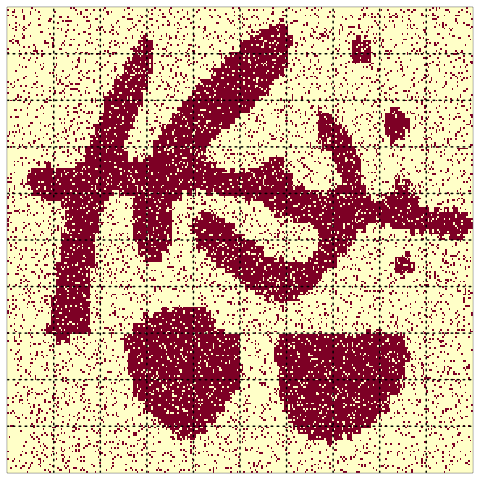}
\label{fig:binsubfig6}}%
\hspace{-0.01\textwidth}
\subfloat[H.\&V.-shift]{
  \includegraphics[width=0.141\textwidth]{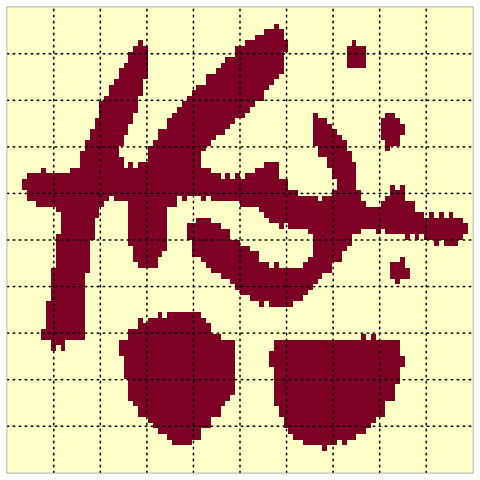}
\label{fig:binsubfig4}}%
\hspace{-0.01\textwidth}
\subfloat[H.\&V.-S\&P]{
  \includegraphics[width=0.141\textwidth]{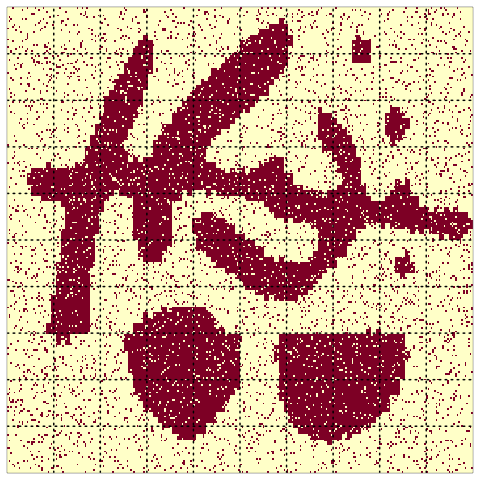}
\label{fig:binsubfig7}}}%
\caption{(a) A binary expanded Besag image to demonstrate the CatSIM metric. (b, d, f) Three spatial translations of the central part of the image. (c, e, g) Images with appropriate salt-and-pepper noise added to match their error rates.}
\label{fig:illustratebinary}
\end{figure*}
\citet{besag1986statistical} presented a $88{\times}100$ hand-drawn
binary scene with intentionally awkward features. We magnified the
image onto a $264{\times} 300$ grid for it to be large enough to run
\begin{figure*}
  \centering
%\begin{minipage}{0.33\textwidth}%
\mbox{\subfloat[True]{\includegraphics[width=.142\textwidth]{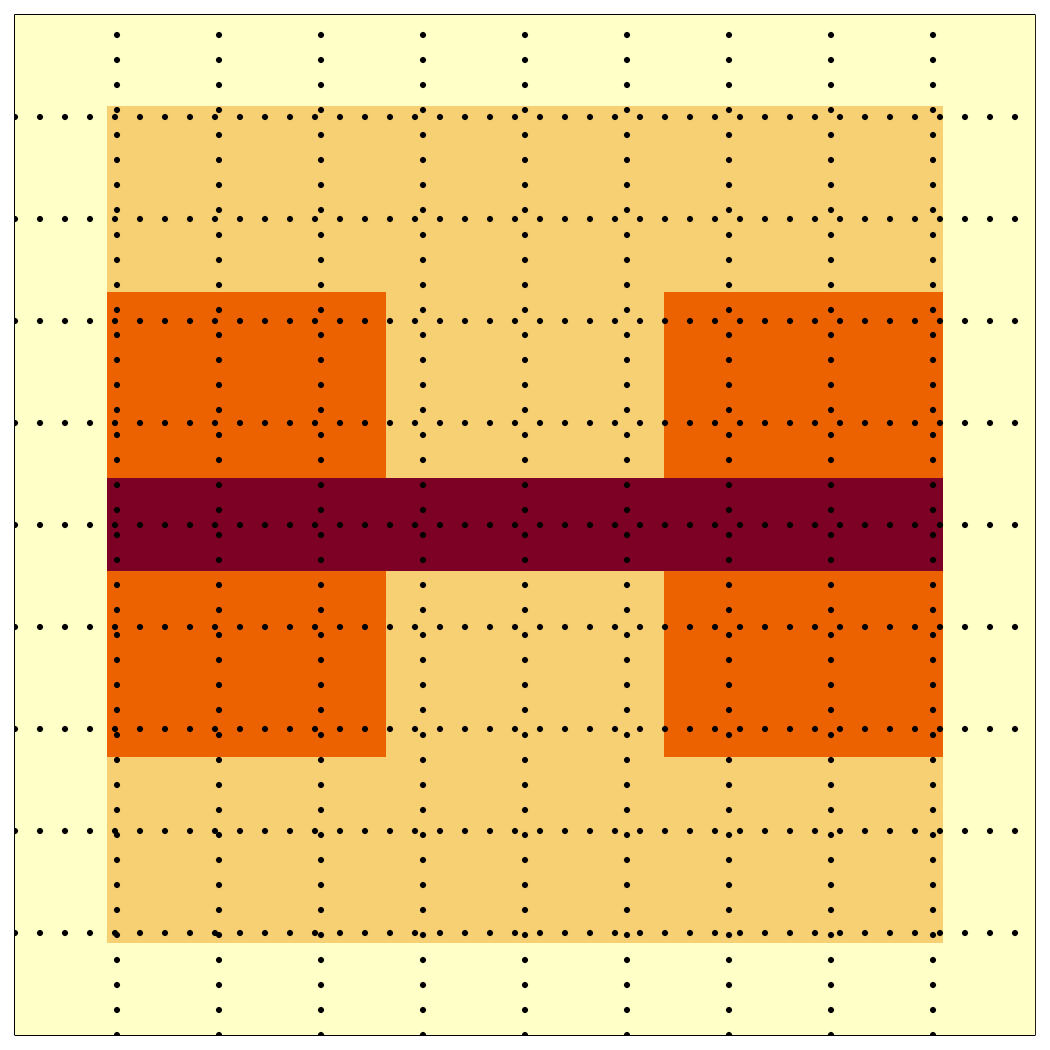}
\label{fig:subfig1}}%
%\end{minipage}%
%\qquad
%\begin{minipage}{0.45\textwidth}%
\hspace{-0.012\textwidth}
\subfloat[H.-shift]{
  \includegraphics[width=0.142\textwidth]{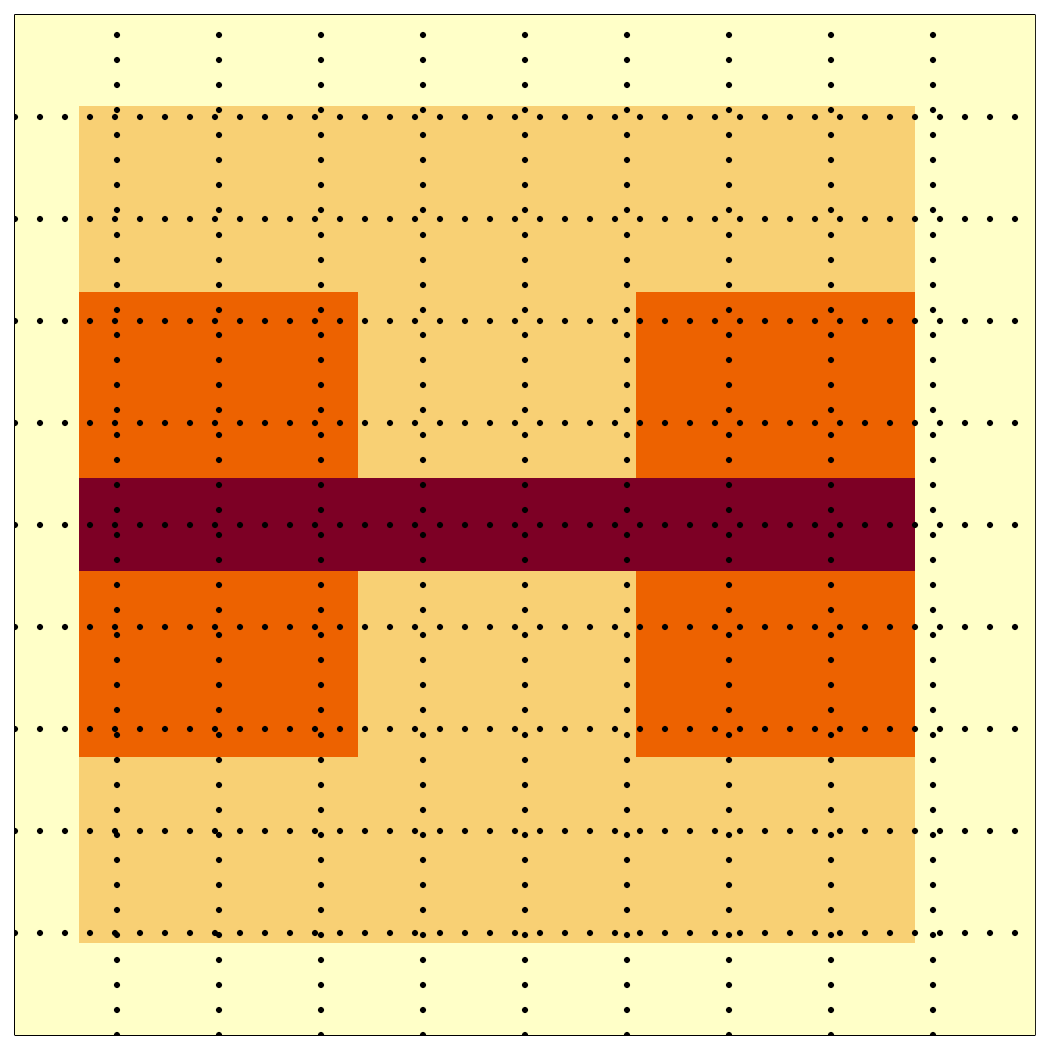}
\label{fig:subfig2}}%
\hspace{-0.012\textwidth}
\subfloat[H.-S\&P]{
  \includegraphics[width=0.142\textwidth]{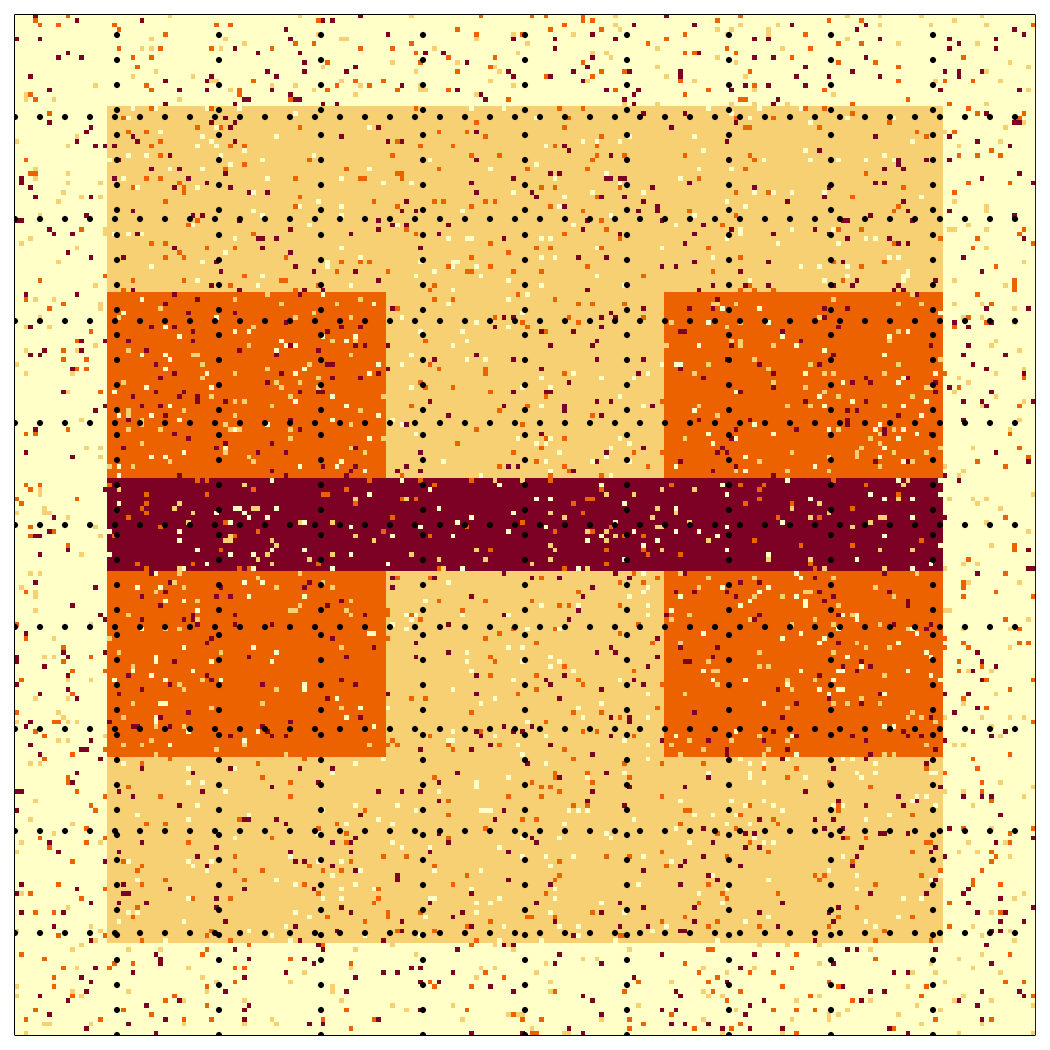}
\label{fig:subfig5}}%
\hspace{-0.012\textwidth}
\subfloat[V.-shift]{
\includegraphics[width=0.142\textwidth]{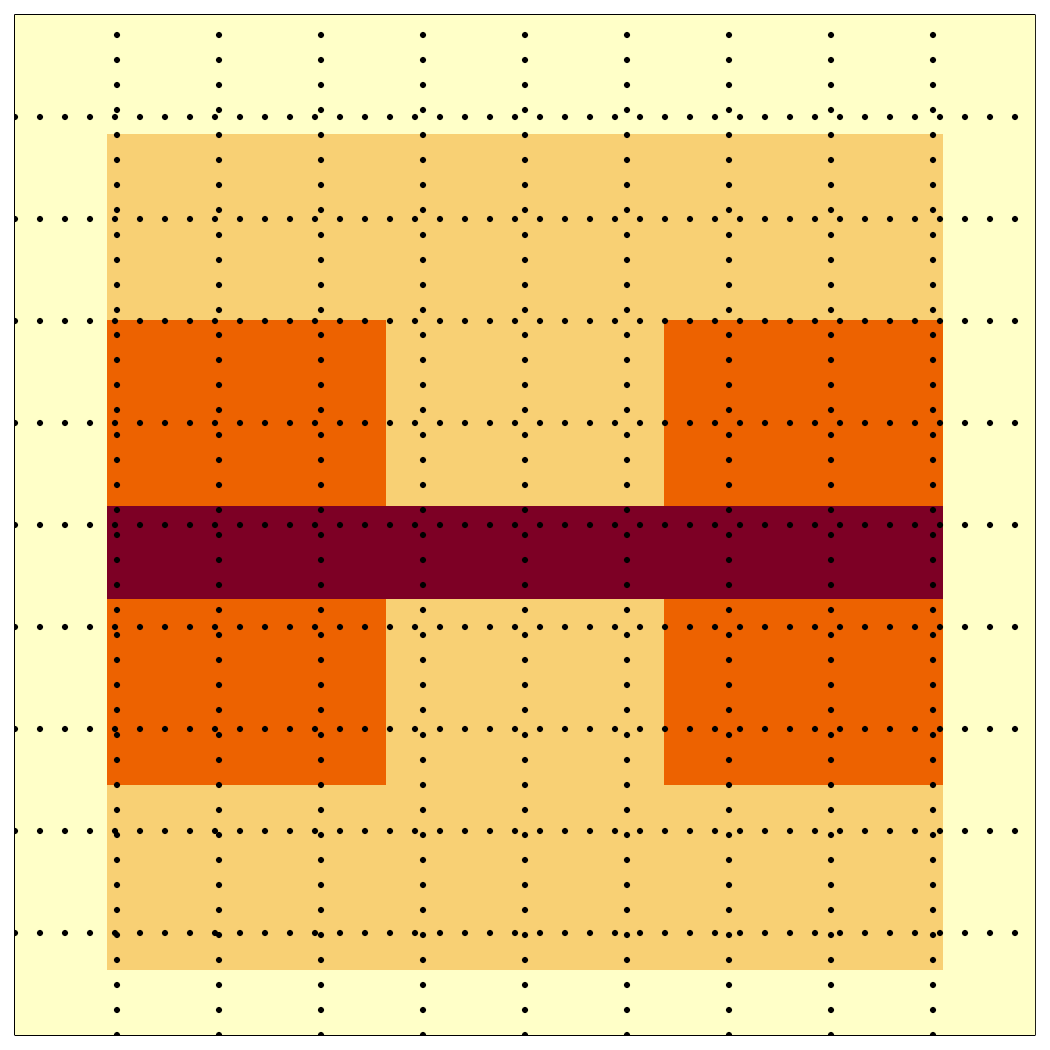}
\label{fig:subfig3}}%
\hspace{-0.012\textwidth}
\subfloat[V.-S\&P]{
  \includegraphics[width=0.142\textwidth]{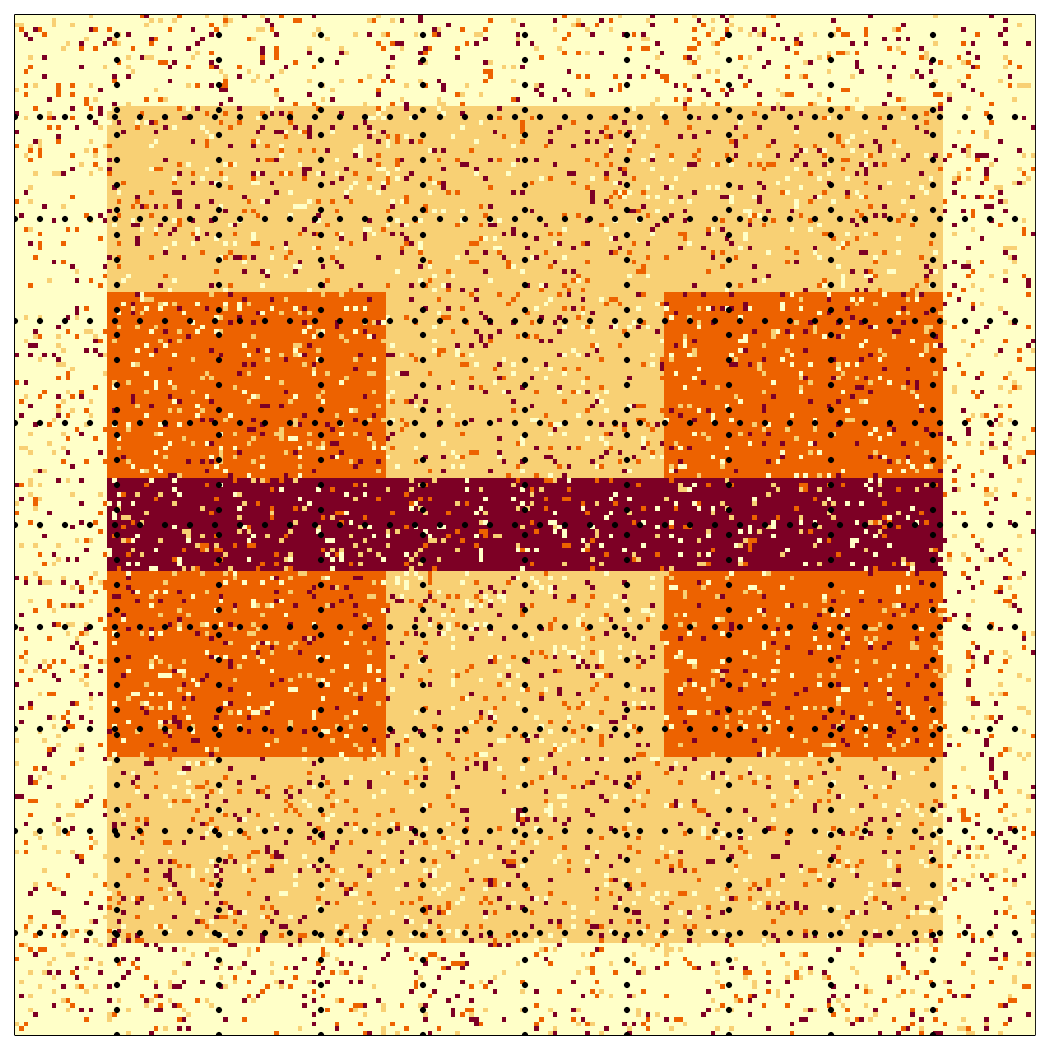}
\label{fig:subfig6}}%
\hspace{-0.012\textwidth}
\subfloat[H.\&V.-shift]{
  \includegraphics[width=0.142\textwidth]{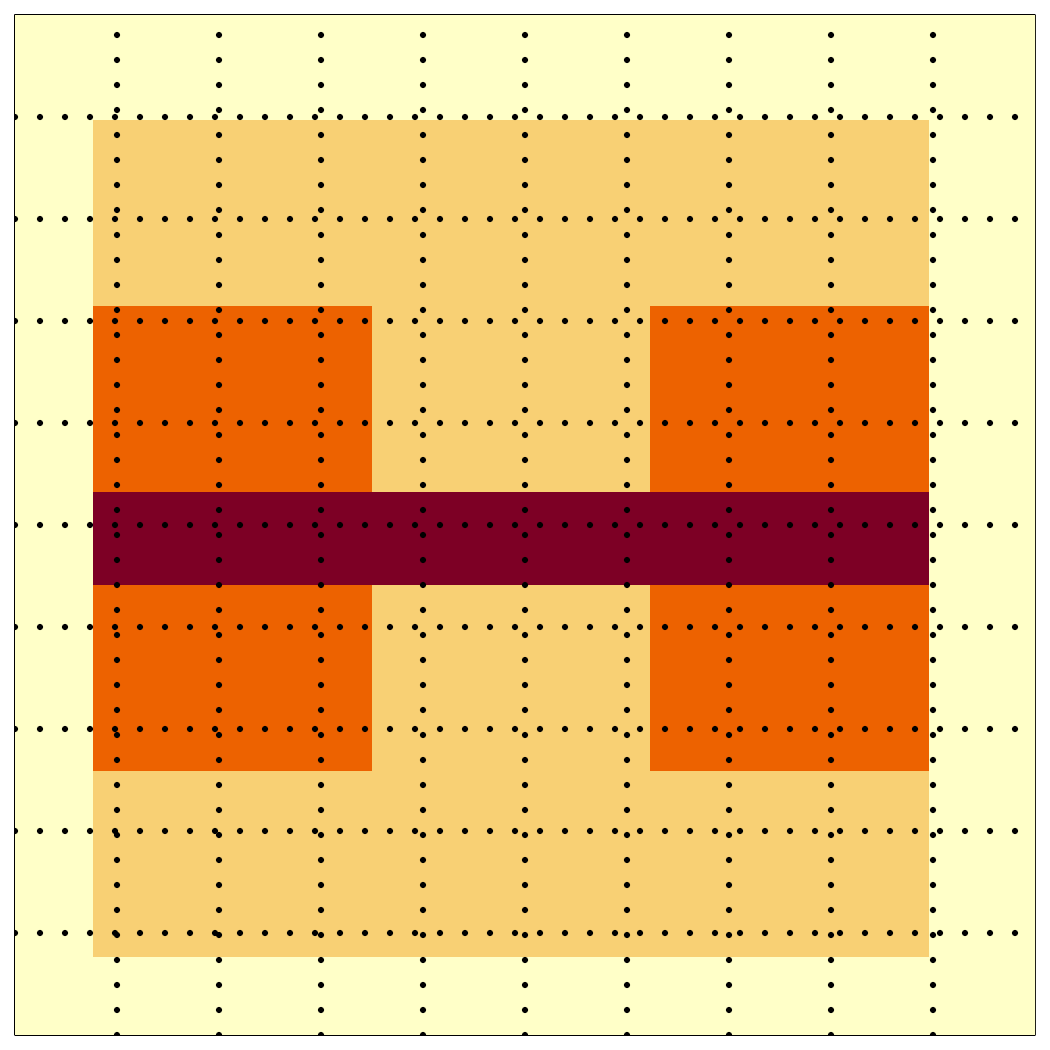}
  \label{fig:subfig4}}%
\hspace{-0.012\textwidth}
\subfloat[H.\&V.-S\&P]{
  \includegraphics[width=0.142\textwidth]{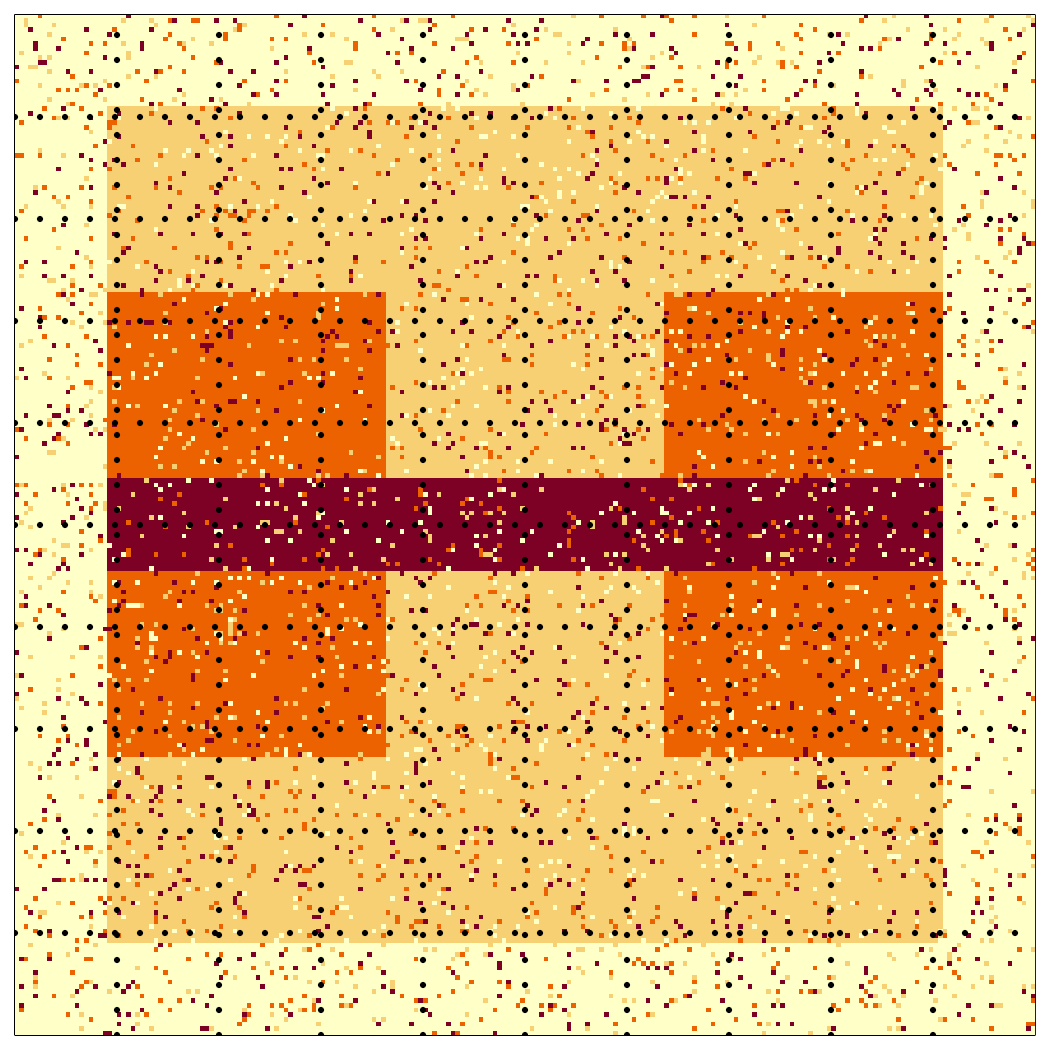}
  \label{fig:subfig7}}}%
\caption{(a) A constructed multi-category image to demonstrate the CatSIM metric. (b, d, f) Three spatial translations of the central part of the image. (c, e, g) Images with appropriate salt-and-pepper noise added to obtain matching error rates.}
\label{fig:illustratemulti}
\end{figure*}
CatSIM ($\kappa$) with five layers and to accommodate spatial translations. This
  expanded Besag image (EBI)~(Figure~\ref{fig:illustratebinary}\subref{fig:binsubfig1}) was trimmed
by 12 pixels off the right and bottom 
margins to allow for  translations of the central portion. We created
horizontally-shifted (by 6 
pixels, Figure~\ref{fig:illustratebinary}\subref{fig:binsubfig2}), vertical-shifted (by 6 pixels, Figure~\ref{fig:illustratebinary}\subref{fig:binsubfig3}) and
horizontally-and-vertically shifted (by 3 pixels in each
direction, Figure~\ref{fig:illustratebinary}\subref{fig:binsubfig4}) versions of the
EBI. Additional degraded
versions~(Figures~\ref{fig:illustratebinary}\subref{fig:binsubfig5},~\ref{fig:illustratebinary}\subref{fig:binsubfig6}, and
\ref{fig:illustratebinary}\subref{fig:binsubfig7}) of the EBI were created by adding 
salt-and-pepper noise (S\&P) with error rates matching those of each of the
shifted images.

  Tables~\ref{tab:besag}~and~\ref{tab:addbinill} (in Appendix~\ref{app:besag})  illustrate the CatSIM,
CW-SSIM as well as  space-agnostic metrics on the different cases of
Figure~\ref{fig:illustratebinary}.  
\begin{table}[ht]
  \centering
  \caption{CatSIM ($\kappa$) and other metrics for different distortions of the
    EBI of Figure~\ref{fig:illustratebinary}.}\label{tab:besag}
   \addtolength{\tabcolsep}{-3pt}
    \begin{tabular}{lcccccc}
      \toprule  
 \parbox[b]{1cm}{Image} & \parbox[t]{.92cm}{CatSIM  5 levels} &  \parbox[t]{.92cm}{CatSIM 1 level} &  \parbox[t]{.91cm}{CatSIM (whole)}  & $\mAR$ & $\kappa$ & CW-SSIM\\
  \midrule
 H Shift         & 0.594 & 0.464 & 0.763 & 0.627 & 0.763 & 0.831 \\%  0.720 &
 H - S \& P      & 0.515 & 0.092 & 0.769 & 0.630 & 0.771 & 0.783\\ %  0.734 &
 V Shift         & 0.569 & 0.449 & 0.751 & 0.610 & 0.751 & 0.752\\ %  0.708 &
 V - S \& P      & 0.516 & 0.090 & 0.756 & 0.613 & 0.759 & 0.780\\ %  0.723 &
 H \& V Shift    & 0.658 & 0.561 & 0.827 & 0.720 & 0.827 & 0.834\\ %  0.788 &
 H \& V - S \& P & 0.557 & 0.110 & 0.832 & 0.725 & 0.834 & 0.810\\ %  0.799 &
   \bottomrule
\end{tabular}
\end{table}
Each pair (shifted and matching S\&P-degraded original) of figures have
similar values for $\mAR$ and $\kappa$. On the other hand, CW-SSIM %has some
                                %separation between the shifted and
                                %noisy versions. For two of the pairs,
                                %the CW-SSIM
rates the horizontal- and horizontal-and-vertically-shifted versions
higher than their matching noisy counterparts but lower for
the vertically-shifted version. However, the translated
images are almost visually indistinguishable from the original, being just minor spatial perturbations, while the noisy versions
are more discordant, and these factors should be reflected in a metric that
mimics the visual system. We compute three different versions of
CatSIM: the (software) default (using Cohen's $\kappa$) with five different scales,  
the default but with only the first scale (no downsampling), and one
computing at once the index for the entire image on one scale (rather
than averaging results from a sliding window). CatSIM rates the
spatially-shifted images differently from the S\&P-degraded images. The difference
is stark when considering only one level, 
to the point that it cannot pick up the structural similarity at that
scale. How much this 
matters in the final index depends on the weights chosen for each level. CatSIM 
calculated over the whole image (rather than a sliding window) is
expectedly not much 
different from $\kappa$, which it is based on. %Further detail
                                %and additional similarity metrics are
                                %in Appendix~\ref{app:illus}.
Interestingly and like 1-level CatSIM, MS-SSIM~(Table~\ref{tab:addbinill}) 
applied to 0-1 images assumed to be in continuous space, 
regards the degraded images very poorly with rates that are %perhaps
not justified visually. In summary, the
5-level CatSIM (with default settings) provides the most consistent representation of the scene.

\subsubsection{A four-class image}\label{subsec:illustrationmulti}
Figure~\ref{fig:subfig1} is our example of a 
$220{\times}220$ four-class image, with several distinct 
spatial regions that are vertically and horizontally
symmetric, around the middle, in each dimension. % The image is both horizontally and vertically
% symmetric.
MS-SSIM and CW-SSIM are not easily extended to the multinary case, so we only study the behavior of CatSIM and the space-unaware
metrics to spatial shifts
(Figures~\ref{fig:illustratemulti}\subref{fig:subfig2},~\ref{fig:illustratemulti}\subref{fig:subfig3}~\ref{fig:illustratemulti}\subref{fig:subfig4})
and matching S\&P degradations (Figures~\ref{fig:illustratemulti}\subref{fig:subfig5},~\ref{fig:illustratemulti}\subref{fig:subfig6}.~\ref{fig:illustratemulti}\subref{fig:subfig7}) of the image obtained in the same manner as in
Section~\ref{subsec:illustrationbin}.
\begin{comment}
As in the previous illustration, we provide three spatial distortions
(moving the central set of objects horizontally, vertically,
and in both directions).
We then generate, using salt and pepper noise, distorted versions of the
original image which match these three in their agreement with the original image, illustrated in
Figure~\ref{fig:illustratemulti}. Values obtained by

As in the previous example, we can compare the performance of the CatSIM ($\kappa$) metric to
other measures of agreement.
\end{comment}

Tables~\ref{tab:illustration} and~\ref{tab:addcatill} 
%we report the default CatSIM metric with five levels
\begin{table}[ht]
  \centering
     \addtolength{\tabcolsep}{-3pt}
     \caption{CatSIM ($\kappa$) and other metrics for different distortions of
       the four-class image of Figure~\ref{fig:illustratemulti}.
      }\label{tab:illustration}
\begin{tabular}{lcccccc}
  \toprule
 Image & \parbox[t]{.95cm}{CatSIM  5 levels} &  \parbox[t]{.95cm}{CatSIM 1 level} &  \parbox[t]{.95cm}{CatSIM (whole)} & $\mAR$ & $\kappa$ \\ 
  \midrule
  H Shift         & 0.816 & 0.686 & 0.906 & 0.828 & 0.906 \\ % 0.936 &
  H - S \& P      & 0.610 & 0.105 & 0.906 & 0.842 & 0.907 \\ % 0.935 &
  V Shift         & 0.652 & 0.462 & 0.827 & 0.723 & 0.827 \\ % 0.881 &
  V - S \& P      & 0.548 & 0.079 & 0.825 & 0.717 & 0.827 \\ % 0.879 &
  H \& V Shift    & 0.814 & 0.533 & 0.869 & 0.777 & 0.869 \\ % 0.910 &
  H \& V - S \& P & 0.565 & 0.093 & 0.874 & 0.790 & 0.875 \\ % 0.913 &
   \bottomrule
\end{tabular}
\end{table}
list the index values.
\begin{comment}
of scaling and a window size of 11, the default CatSIM index with with only one level (i.e., no
downsampling), the CatSIM index computed over the entire image (i.e., a window size of 220 pixels),
the Adjusted Rand index, and Cohen's $\kappa$.
While the Adjusted Rand ($\mAR$) and Cohen's $\kappa$ were approximately the same for all
the pairs of images, the five-level CatSIM index was able to distinguish the spatially shifted
images while still indicating the images were similar. The single-level CatSIM and the whole-image CatSIM
have the same issues as in the binary image: the single level is unable to capture the similarity
between the noisy image and the whole-image version is unable to distinguish the two images.
More complete results are included in Appendix~\ref{app:illus}.
\end{comment}
  % latex table generated in R 3.6.2 by xtable 1.8-4 package
  % Mon Mar  9 13:57:42 2020
As in the EBI example, $\mAR$ and $\kappa$ values are similar for each
pair of the shifted and its S\&P-matched twin. The case for
the whole-image CatSIM values is similar. However, the default 5-level and
1-level CatSIM indices clearly distinguish between images that are
minor spatial perturbations over images degraded with added
noise. The 1-level CatSIM index is unnecessarily harsh on the
S\&P-degraded images, being very poor at recognizing similarity with
the original. The 5-layer version (with default parameters) recovers
this similarity because the downsampling smooths out the added noise.
%Treating the entire image at once in CatSIM captures the similarity
%between the distorted images and the original but does not
%distinguish well between the methods of distortion; it behaves
%similarly to the Adjusted Rand and Cohen's $\kappa$ metrics.

\subsubsection{A highly-imbalanced binary image}\label{subsec:phantom}
\begin{figure}[h]
%  \centering
  \vspace{-0.02\textheight}
  \begin{minipage}{0.97\columnwidth}
\hspace{-0.025\textwidth}
\subfloat[Truth]{
\includegraphics[width=.19\linewidth]{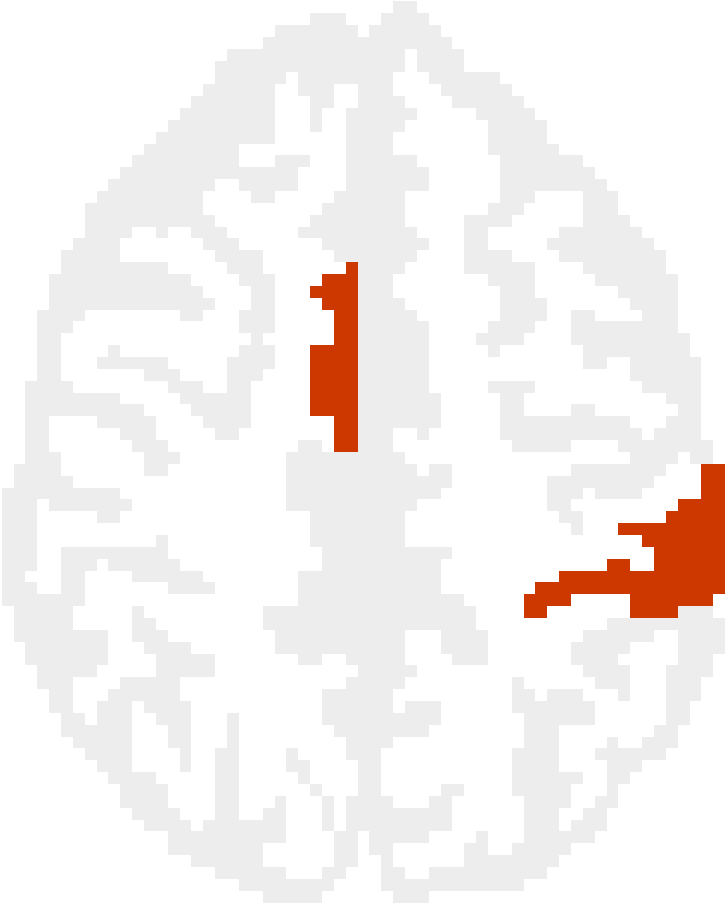}  
\label{fig:hoff-first}}%
~%
\hspace{-0.02\textwidth}
\subfloat[D+1]{
\includegraphics[width=.19\linewidth]{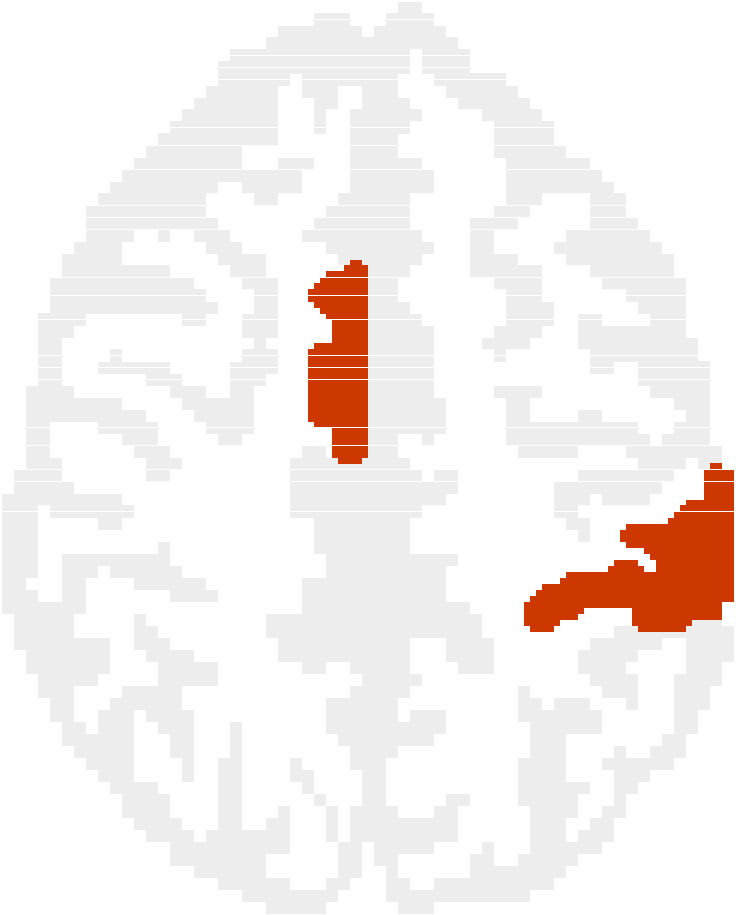}  
\label{fig:hoff-expand-01}}%
~%
\hspace{-0.02\textwidth}
\subfloat[D+2]{
\includegraphics[width=.19\linewidth]{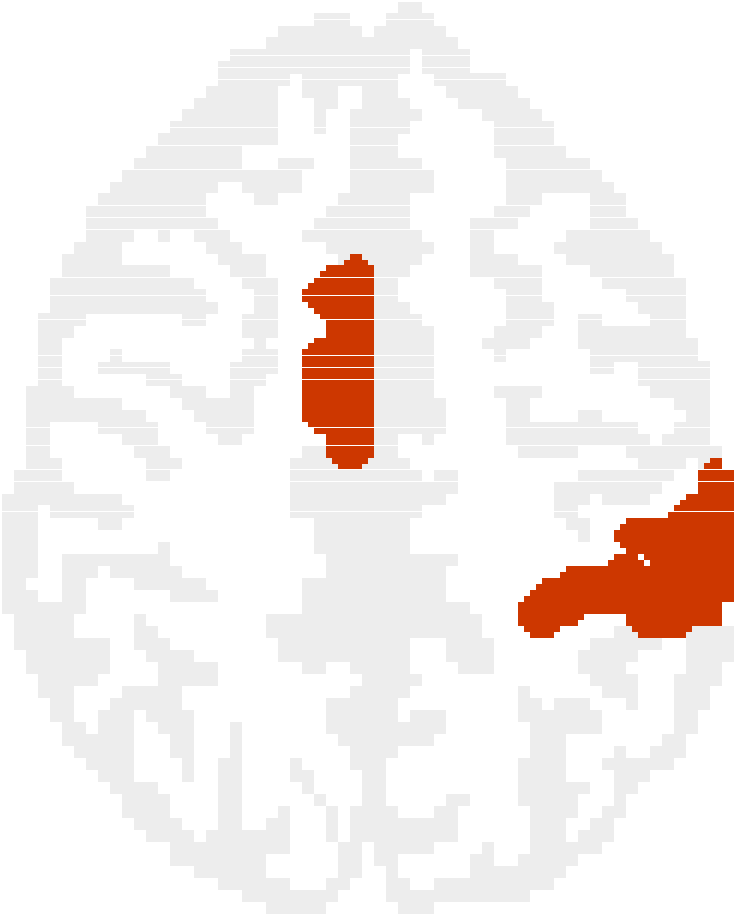}  
\label{fig:hoff-second}}%
~
% \hspace{-0.02\textwidth}
\subfloat[E-1]{
\includegraphics[width=.19\linewidth]{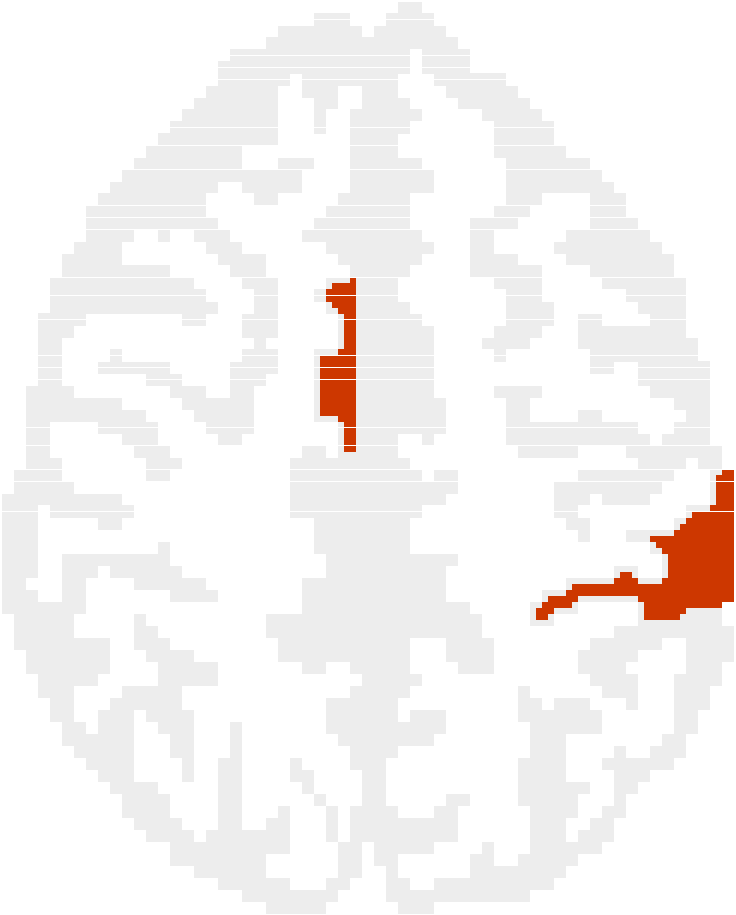}  
\label{fig:hoff-contract-01}}%
~%
\hspace{-0.02\textwidth}
\subfloat[E-2]{
\includegraphics[width=.19\linewidth]{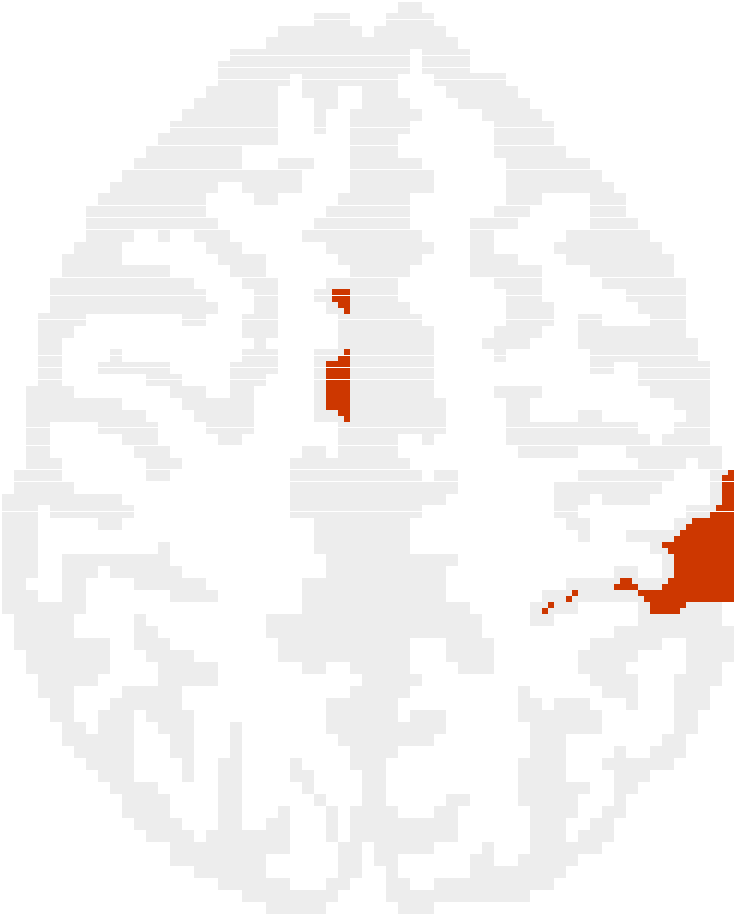}  
\label{fig:hoff-third}}%
\\%
\hspace{-0.025\textwidth}
\subfloat[S$\uparrow$1]{
\includegraphics[width=.19\linewidth]{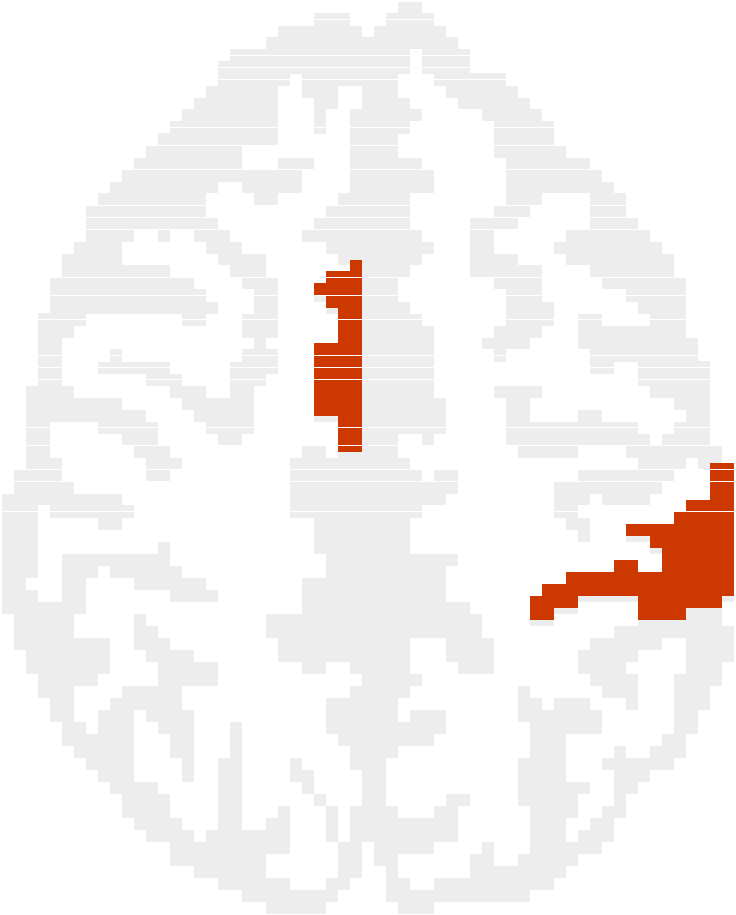}  
\label{fig:hoff-shiftup}}%
~%
\hspace{-0.02\textwidth}
\subfloat[S$\downarrow$2]{
\includegraphics[width=.19\linewidth]{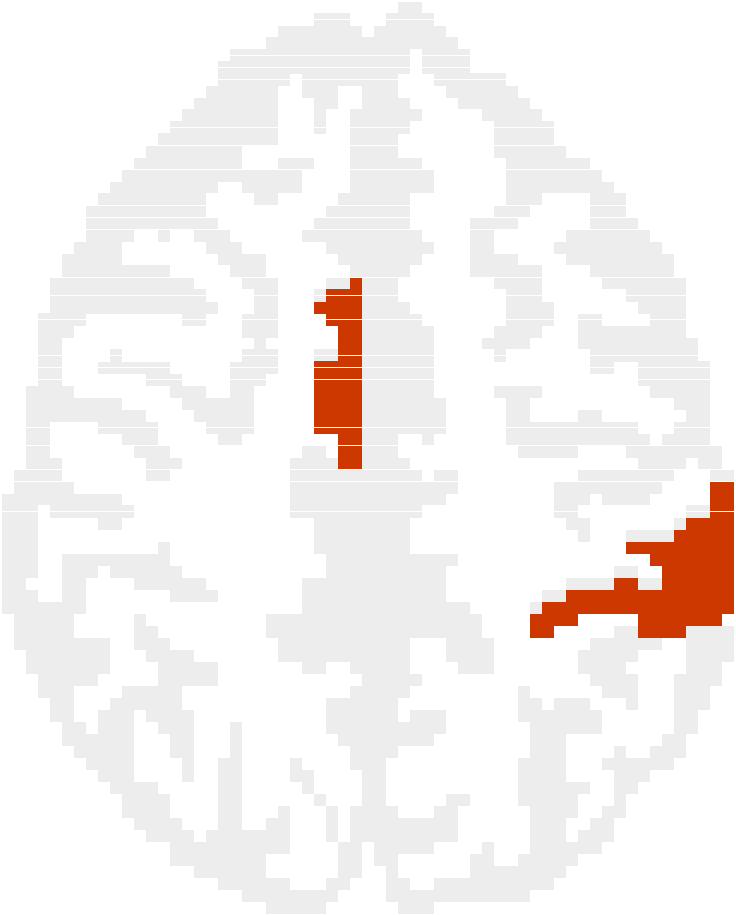}  
\label{fig:hoff-shiftdown}}%
~%
\hspace{-0.02\textwidth}
\subfloat[+$\frac{\varepsilon}{100}$]{
\includegraphics[width=.19\linewidth]{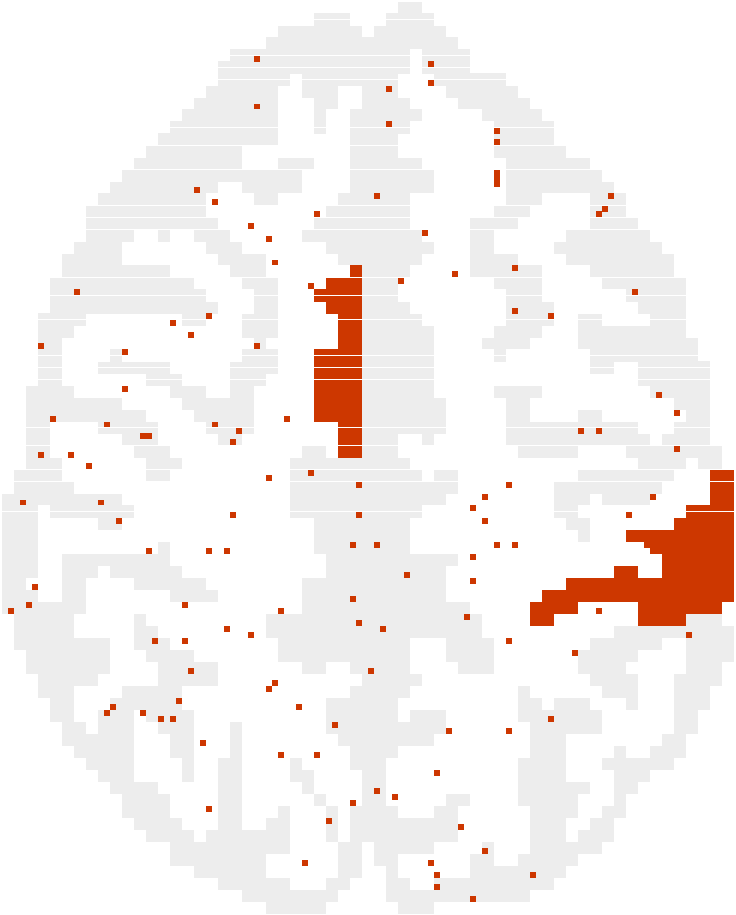}  
\label{fig:hoff-noise-01}}%
~%
\hspace{-0.02\textwidth}
\subfloat[+$\frac{3\varepsilon}{100}$]{
\includegraphics[width=.19\linewidth]{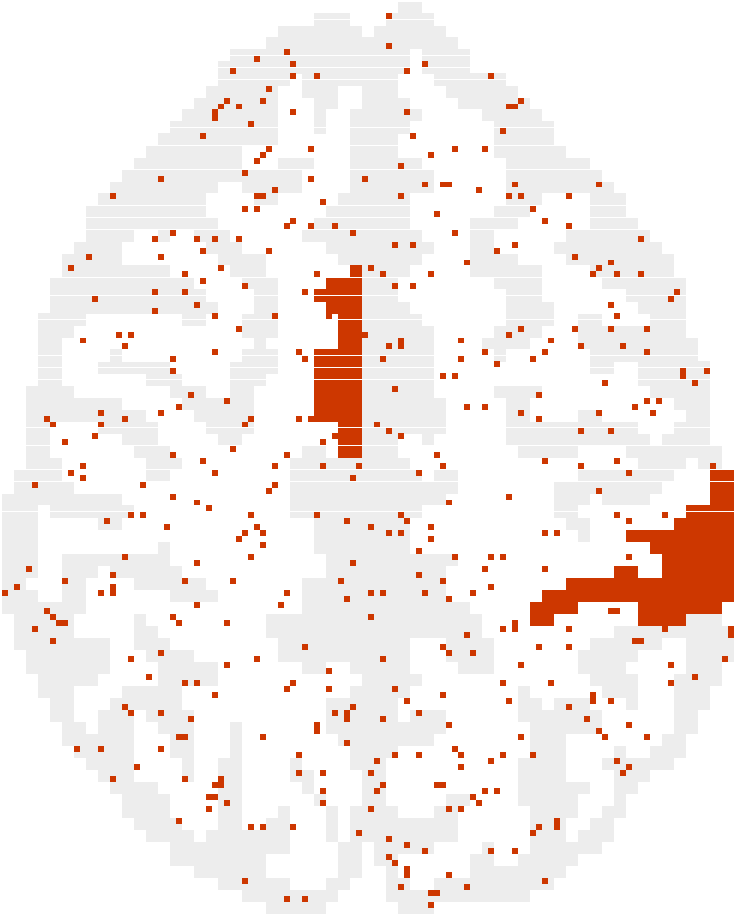}  
\label{fig:hoff-noise-03}}%
~%
\hspace{-0.02\textwidth}
\subfloat[S$\uparrow$1+$\frac{2\varepsilon}{100}$]{
\includegraphics[width=.19\linewidth]{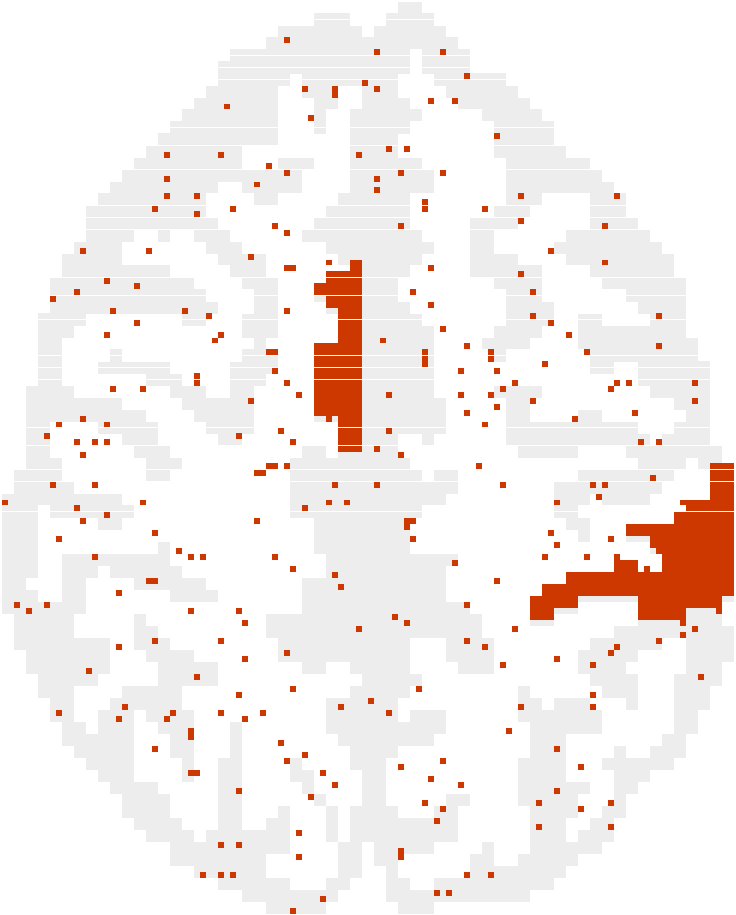}  
\label{fig:hoff-fourth}}%
\end{minipage}
%\caption{The in-brain portion of the ground truth modified Hoffman
%phantom   (Figure~\ref{fig:hoff-first}) and its dilation
%(Figures~\ref{fig:hoff-expand-01},~\ref{fig:hoff-second}),   erosion
%(Figures~\ref{fig:hoff-contract-01},~\ref{fig:hoff-third}),
%shifting up (Figure~\ref{fig:hoff-shiftup}) and down
%(Figure~\ref{fig:hoff-shiftdown}),   adding 1\% and 3\% noise
%(Figures~\ref{fig:hoff-noise-01},~\ref{fig:hoff-noise-03}), and shifting with the addition of 2\% noise (Figure~\ref{fig:hoff-fourth}).}\label{fig:hoffman}
\caption{(a) The in-brain portion of the modified Hoffman activation   phantom of \citet{almodovarandmaitra19} 
and distortions of its activated  (dark-shaded) regions: dilation   by (b) 1 and (c) 2 pixels, erosion by (d) 1 and (e) 2 pixels, (f) upward shift by 1 pixel, (g) downward shift by 2 pixels. Other distortions add noise by randomly activating (h) 1\%, (i) 3\% and (j) 2\% inactivated in-brain pixels, the last after also shifting the activated region upward by 1 pixel. 
%(Figure~\ref{fig:hoff-shiftdown}),   adding 1\% and 3\% noise
%(Figures~\ref{fig:hoff-noise-01},~\ref{fig:hoff-noise-03}), and
%shifting with the addition of 2\% noise
%(Figure~\ref{fig:hoff-fourth}).
}
\label{fig:hoffman}
\end{figure}

Our next illustration is on a binary image with a mask and with
disproportionate class sizes, as happens, say, in activation detection
with functional Magnetic Resonance Imaging (fMRI)
where only about 2-3\% of voxels are expected to be activated~\citep{chenandmaitra23}. Our ground
truth, in~Figure~\ref{fig:hoffman}\subref{fig:hoff-first}, is a $256{\times}256$
version of the modified $128{\times}128$ Hoffman activation phantom~\citep{Hoffman1990} with a small proportion (3.98\%) of  truly activated in-brain pixels~\citep{almodovarandmaitra19}. The in-brain pixels form a mask, which also renders  methods such as CW-SSIM inapplicable. 

We prepared nine distorted versions of the image using mathematical
morphology and other operations. Specifically, we dilated  
the activated regions by one (D+1) and two (D+2) pixels (with one or two
inactivated pixels  orthogonally adjacent to an activated pixel changing status),
eroded the activated regions by one (E-1) and two (E-2) pixels
(essentially deactivating any activated pixel  orthogonally adjacent to an
inactivated one), the activation regions shifted up one (S$\uparrow$1)
and down two (S$\downarrow$2) pixels, the original image with 1\%
(+$\frac\varepsilon{100}$) and 3\% (+$\frac{3\varepsilon}{100}$) of the inactive pixels 
randomly activated, and the original image with the activated regions shifted up
by one pixel and adding 2\% noise (S$\uparrow$1+2\%).
%in~\ref{fig:hoffapp}.

% latex table generated in R 3.6.3 by xtable 1.8-4 package
% Sat Apr  4 00:47:11 2020
The similarity indices for these distorted images~(Table~\ref{tab:hoffmanacc})
\begin{table}[ht]
  \centering
  \caption{Similarity indices for the distorted and degraded images of Figure~\ref{fig:hoffman}.}\label{tab:hoffmanacc}
\begin{tabular}{lrrrcc}
  \toprule
  & $\mJ$ & Dice & Acc. &  \parbox[t]{.92cm}{CatSIM ($\mJ$)} & \parbox[t]{.92cm}{CatSIM ($\kappa$)} \\ 
  \midrule
Dilated (D+1)& 0.75 & 0.86 & 0.99 & 0.74 & 0.94 \\ 
Dilated (D+2) & 0.61 & 0.76 & 0.97 & 0.59 & 0.88 \\ 
Eroded (E-1)& 0.68 & 0.81 & 0.99 & 0.61 & 0.90 \\ 
Eroded (E-2) & 0.43 & 0.60 & 0.98 & 0.33 & 0.72 \\ 
  Shift Up (S$\uparrow$1) & 0.83 & 0.91 & 0.99 & 0.75 & 0.94 \\ 
Shift Down (S$\downarrow$2) & 0.68 & 0.81 & 0.99 & 0.52 & 0.77 \\ 
Noise (+$\frac{\varepsilon}{100}$) & 0.81 & 0.90 & 0.99 & 0.63 & 0.80 \\ 
Noise (+$\frac{3\varepsilon}{100}$) & 0.57 & 0.73 & 0.97 & 0.45 & 0.58 \\ 
Shift+Noise (S$\uparrow$1+$\frac{2\varepsilon}{100}$) & 0.58 & 0.73 & 0.97 & 0.42 & 0.63 \\ 
   \bottomrule
\end{tabular}
\end{table}
all agree with our expectation that an increase in distortion - a larger shift
or more noise - will have a lower similarity to the base image. Accuracy (Acc.) 
is a poor measure in this context because the feature that is important to capture here is
the difference in activation, and it reports a very high agreement for all
of the distorted images (because the pixels in all the images are 
mostly of one class). The Jaccard and Dice indices deteriorate as 
desired with increasing distortion and the CatSIM methods, both with $\mJ$ or with Cohen's $\kappa$, that is, CatSIM($\mJ$) or CatSIM($\kappa$), do well, even though the latter is not specifically designed for this behavior and does not capture the difference as well 
as the other indices. Compared to $\mJ$, CatSIM($\mJ$) penalizes noise
in the images more than for minor perturbations that do not affect the basic
spatial extent of the activated region. We now study the impact of the
(five) layers in the calculation of CatSIM($\mJ$).

Table~\ref{tab:catlayers} illustrates the utility of the five layers
in the calculation of the index.
\begin{table}[ht]
  \centering
  \caption{CatSIM ($\mJ$) values for each layer for different types of distortions.
    Layer 1 is the image itself while subsequent layers downsample the
    image in the previous layer by a factor of 2.
    }\label{tab:catlayers}
\begin{tabular}{lccccc}
  \toprule
 & Layer 1 &  2 &  3 & 4 &  5 \\ 
  \midrule
Dilated (D+1) & 0.60 & 0.65 & 0.68 & 0.83 & 1.00 \\ 
Dilated (D+2) & 0.43 & 0.48 & 0.53 & 0.62 & 1.00 \\ 
Eroded (E-1) & 0.52 & 0.51 & 0.57 & 0.56 & 1.00 \\ 
Eroded (E-2) & 0.28 & 0.32 & 0.34 & 0.38 & 0.35 \\ 
Shift Up (S$\uparrow$1) & 0.72 & 0.59 & 0.72 & 0.75 & 1.00 \\ 
Shift Down (S$\downarrow$2) & 0.55 & 0.59 & 0.58 & 0.59 & 0.35 \\ 
Noise (+$\frac{\varepsilon}{100}$) & 0.12 & 0.86 & 1.00 & 1.00 & 1.00 \\ 
Noise (+$\frac{3\varepsilon}{100}$) & 0.04 & 0.46 & 1.00 & 1.00 & 1.00 \\ 
Shift+Noise  (S$\uparrow$1+$\frac{2\varepsilon}{100}$) & 0.05 & 0.48 & 0.79 & 0.75 & 1.00 \\ 
   \bottomrule
\end{tabular}
\end{table}
The first level, without any downsampling, rates any added
noise poorly, but higher levels smooth out that difference. Relatively
large differences, such as caused by double erosion (E-2), damage the
image's rating across all scales. Any scale larger than this has ratings of either 1 or 0 in this application as
the activated class disappears completely.
The optimal number of levels depends on the size of the
features in the image, the type of distortion, and
the demands of the application, but using five equally weighted
levels seems, in this application and, in our experience, more generally, to strike a balance 
and provide good  performance.

\subsubsection{Impact of Weight Selection}\label{subsec:weights}

\begin{table*}[ht]
  \centering
  \caption{CatSIM ($\kappa$) and CatSIM ($\mAR$) metrics for different distortions of (left block) the    EBI of Figure~\ref{fig:illustratebinary} and (right block) 
the multicategory image of Figure~\ref{fig:illustratemulti} using a flat weighting scheme, a linear increasing
    weighting scheme, and a linear decreasing weighting scheme.}\label{tab:both}
\begin{tabular}{l|cccccc||cccccc}
  \toprule
   & \multicolumn{6}{c||}{\bf Extended Besag image (EBI)}&\multicolumn{6}{c}{\bf Multicategory image}\\
  \parbox[b]{1.2cm}{Image Operation} & \parbox[t]{.81cm}{CatSIM ($\kappa$)} & \parbox[t]{.81cm}{CatSIM ($\mAR$)} & \parbox[t]{.88cm}{CatSIM ($\kappa$) $\uparrow$ } & \parbox[t]{.9cm}{CatSIM ($\mAR$)  $\uparrow$ }& \parbox[t]{.88cm}{CatSIM ($\kappa$)  $\downarrow$} & \parbox[t]{.9cm}{CatSIM ($\mAR$)  $\downarrow$}
 & \parbox[t]{.81cm}{CatSIM ($\kappa$)} & \parbox[t]{.81cm}{CatSIM ($\mAR$)} & \parbox[t]{.88cm}{CatSIM ($\kappa$) $\uparrow$ } & \parbox[t]{.9cm}{CatSIM ($\mAR$)  $\uparrow$ }& \parbox[t]{.88cm}{CatSIM ($\kappa$)  $\downarrow$} & \parbox[t]{.9cm}{CatSIM ($\mAR$)  $\downarrow$}  \\ 
  \midrule
   H Shift   & 0.594 & 0.734 & 0.640 & 0.749 & 0.551 & 0.719 & 0.816 & 0.879 & 0.854 & 0.905 & 0.779 & 0.854    \\ 
   H - S \& P & 0.515 & 0.723 & 0.738 & 0.855 & 0.359 & 0.612 & 0.610 & 0.799 & 0.835 & 0.923 & 0.445 & 0.692  \\ 
  V Shift & 0.569 & 0.714 & 0.602 & 0.720 & 0.537 & 0.708 & 0.652 & 0.770 & 0.718 & 0.817 & 0.591 & 0.727 \\ 
  V - S \& P  & 0.503 & 0.715 & 0.727 & 0.849 & 0.348 & 0.603 & 0.533 & 0.738 & 0.776 & 0.887 & 0.366 & 0.614  \\ 
  H \& V Shift & 0.658 & 0.771 & 0.682 & 0.770 & 0.635 & 0.772 & 0.814 & 0.886 & 0.902 & 0.942 & 0.736 & 0.834  \\ 
  H \& V, S \& P & 0.557 & 0.758 & 0.775 & 0.881 & 0.401 & 0.651 & 0.577 & 0.773 & 0.811 & 0.909 & 0.410 & 0.657  \\ 
   \bottomrule
\end{tabular}
\end{table*}
The analysis in Section~\ref{subsec:phantom} suggests potential sensitivity of the metric to the choice of weighting. The analysis in the last section was done with
equal weighting on each level (with weights adding to unity).
We examine two other weighting schemes: an increasing and a decreasing
linear ramp (weights of $1/15, 2/15, \ldots, 5/15$ in increasing and
decreasing order) for both the default CatSIM($\kappa$) and CatSIM($\mAR$).

% latex table generated in R 4.0.5 by xtable 1.8-4 package
% Sat Oct 23 22:51:12 2021
\begin{comment}
\begin{table}[ht]
  \centering
  \caption{CatSIM ($\kappa$) and CatSIM ($\mAR$) metrics for different distortions of the
    EBI of Figure~\ref{fig:illustratebinary} using a flat weighting scheme, a linear increasing
    weighting scheme, and a linear decreasing weighting scheme.}\label{tab:weightbesag}
\begin{tabular}{lrrrrrr||}
  \toprule
 \parbox[b]{.9cm}{Image} & \parbox[t]{.81cm}{CatSIM ($\kappa$)} & \parbox[t]{.81cm}{CatSIM ($\mAR$)} & \parbox[t]{.88cm}{CatSIM ($\kappa$) $\uparrow$ } & \parbox[t]{.9cm}{CatSIM ($\mAR$)  $\uparrow$ }& \parbox[t]{.88cm}{CatSIM ($\kappa$)  $\downarrow$} & \parbox[t]{.95cm}{CatSIM ($\mAR$)  $\downarrow$} \\ 
  \midrule
   H Shift   & 0.594 & 0.734 & 0.640 & 0.749 & 0.551 & 0.719 \\ 
   H - S \& P & 0.515 & 0.723 & 0.738 & 0.855 & 0.359 & 0.612 \\ 
  V Shift & 0.569 & 0.714 & 0.602 & 0.720 & 0.537 & 0.708 \\ 
  V - S \& P  & 0.503 & 0.715 & 0.727 & 0.849 & 0.348 & 0.603 \\ 
  H \& V Shift & 0.658 & 0.771 & 0.682 & 0.770 & 0.635 & 0.772 \\ 
  H \& V, S \& P & 0.557 & 0.758 & 0.775 & 0.881 & 0.401 & 0.651 \\ 
   \bottomrule
\end{tabular}
\end{table}
\end{comment}
Table~\ref{tab:both} (left block) shows that for the EBI of Figure~\ref{fig:illustratebinary}, the increasing weight scheme has higher ratings in some images, and somewhat larger error rates. For instance, the horizontally shifted
image has a lower rating than the horizontally shifted image with salt and pepper noise added. A similar
problem occurs with the vertically shifted image. However, the decreasing weight scheme does not have this
problem. Indeed, we see that the CatSIM versions with decreasing weights penalize the accumulation  of error slightly
more than the versions with level weights and rate each image slightly lower than their counterparts.

% latex table generated in R 4.0.5 by xtable 1.8-4 package
% Sat Oct 23 23:03:10 2021
\begin{comment}
\begin{table}[ht]
\centering
 \caption{CatSIM ($\kappa$) and CatSIM ($\mAR$) metrics for different distortions of the
    multicategory image of Figure~\ref{fig:illustratemulti} using a flat weighting scheme, a linear increasing
    weighting scheme, and a linear decreasing weighting scheme.}\label{tab:weightmulti}
\begin{tabular}{lrrrrrr}
  \toprule
 \parbox[b]{.9cm}{Image} & \parbox[t]{.81cm}{CatSIM ($\kappa$)} & \parbox[t]{.81cm}{CatSIM ($\mAR$)} & \parbox[t]{.88cm}{CatSIM ($\kappa$) $\uparrow$ } & \parbox[t]{.88cm}{CatSIM ($\mAR$)  $\uparrow$ }& \parbox[t]{.88cm}{CatSIM ($\kappa$)  $\downarrow$} & \parbox[t]{.88cm}{CatSIM ($\mAR$)  $\downarrow$} \\ 
  \midrule
   H Shift & 0.816 & 0.879 & 0.854 & 0.905 & 0.779 & 0.854 \\ 
   H - S \& P & 0.610 & 0.799 & 0.835 & 0.923 & 0.445 & 0.692 \\ 
   V Shift & 0.652 & 0.770 & 0.718 & 0.817 & 0.591 & 0.727 \\ 
   V - S \& P   & 0.533 & 0.738 & 0.776 & 0.887 & 0.366 & 0.614 \\ 
  H \& V Shift & 0.814 & 0.886 & 0.902 & 0.942 & 0.736 & 0.834 \\ 
   H \& V - S \& P & 0.577 & 0.773 & 0.811 & 0.909 & 0.410 & 0.657 \\ 
   \bottomrule
\end{tabular}
\end{table}
\end{comment}
At the same time, Table~\ref{tab:both} (right block) shows similar patterns in the case of the multicategory image from Section~\ref{subsec:illustrationmulti}, albeit to a lesser extent. There are several cases where the addition of noise increases the performance
of a CatSIM index when the weights are higher for higher levels of downsampling. The behavior of CatSIM ($\kappa$)
and CatSIM ($\mAR$) differ with the addition of noise in the case of decreasing weights -- the one with
Cohen's $\kappa$ decreases sharply with the addition of the noise while the one based on the Adjusted Rand has a more muted
response than either level-weighted version. In general, weighting schemes can be chosen by users based on their preferred metric, context or application, but the results here suggest that level weights are a balanced choice which respond
as expected to the addition of noise.
%\subsubsection{Impact of other parameters} We conclude our analysis and discussion here  by noting that we also looked at the choices of the number of levels as well as of $N_u,N_v$ (and $N_w$) but our investigations did not reveal a clear pattern with these choices that could be exploited beyond the default recommendations. 

%%% Local Variables:
%%% mode: latex
%%% TeX-master: "catsim-paper"
%%% End:

\subsection{Image Quality Assessment Surveys}

% \begin{figure*}
%   \centering
%   \includegraphics{figure/images-2-tile.pdf}
%  \caption{The remaining undistorted images from the first survey.}\label{fig:settwo}
% \end{figure*}
Having illustrated CatSIM's ability to capture structural similarity in
categorical images, we now evaluate its ability to represent human
visual perception. % judge the performance of the proposed metric,
We conducted two separate surveys to compare the metric's assessment on binary
and multinary images to that obtained by human judgment of the image quality. We now
discuss the surveys and their results.
\begin{comment}
A mean opinion score (MOS) for each image is constructed by
computing the mean value assigned by the raters of the quality of the image.
The binary images for this test were either generated using random processes, produced
by thresholding grayscale images, or selected from existing binary image data sets.
With the binary images, it is possible to compare perceived
quality to MS-SSIM, PSNR, and some binary measures of the
quality in addition to the various options for CatSIM. For the
multicategory images, comparisons can only be made to metrics which
depend on pixel-wise agreement and disagreement between the two images.

\begin{figure}
  \includegraphics[width=\columnwidth]{figure/gray-4x.pdf}
  \caption{The 12 undistorted $256\times256$ binary images used in the
    survey.}\label{fig:setone}
\end{figure}
%montage image-1.png image-6.png image-7.png image-12.png image-2.png image-3.png image-4.png image-5.png image-8.png image-9.png image-10.png image-11.png -tile x3 -geometry +0+0 -borderwidth 3 -bordercolor lightblue gray-4x.pdf
\end{comment}
\subsubsection{Assessment on Binary Images}
\label{sec:bin}
\begin{figure}[h]
  \includegraphics[width=\columnwidth]{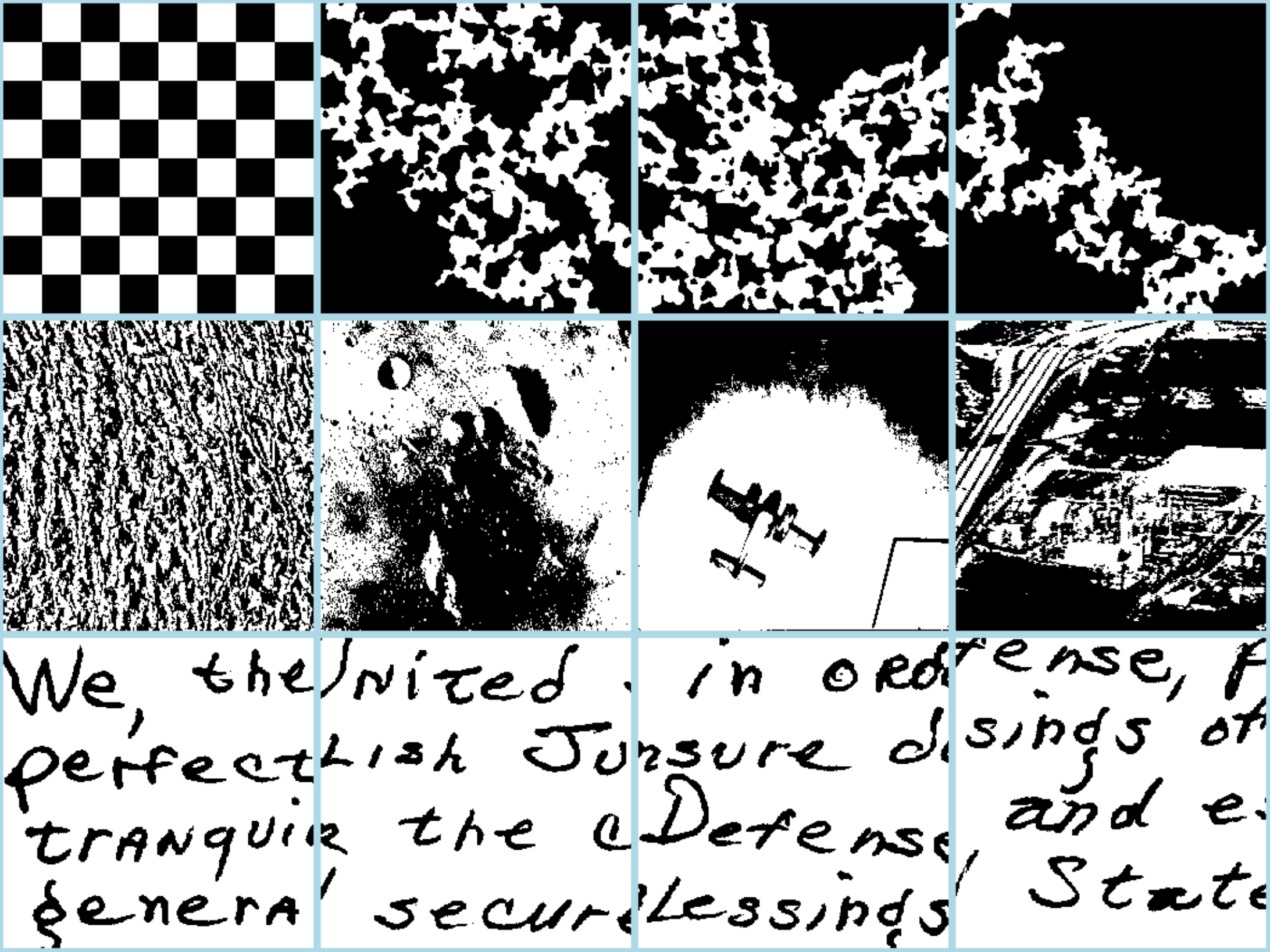}
  \caption{The 12 undistorted $256 \times 256$ binary images from the first survey.}\label{fig:setone}
\end{figure}

% to be inserted once ready
Figure~\ref{fig:setone} displays the twelve  $256{\times}256$ binary
images that, along with their distorted versions, were used in the
survey. The first  image (Figure~\ref{fig:setone}, first
row) is of an 8-squares checker-board pattern, and is followed by
images of three thresholded Gaussian processes. 
The next row of images (Figure~\ref{fig:setone}) were obtained
by thresholding monochrome images (of a texture, the lunar
surface, an aerial view of, first, an airfield with an airplane, and
then a highway overpass) from the
\href{http://sipi.usc.edu/database/database.php}{USC-SIPI image
  database}. The last set of binary images had thresholded versions
of four handwriting samples from the NIST Special Database
19~\citep{NISTdb}. Seventy-four %informed-consenting
adult volunteers
were shown 30 pairs of images, with each pair comprising a randomly chosen
image from  Figure~\ref{fig:setone} and a distorted version. They 
were asked to rate the quality of each distorted image on a scale from
1 to 100, with 100 indicating perfect fidelity. For each of $12{\times}12{=}144$
ground truth-distorted image pairs, we
calculated the mean opinion scores (MOS) over all the respondents who were 
shown that particular image pair. For these image pairs, we compared
the MOS to CW-SSIM, the space-unaware metrics of $\kappa$, $\mAR$, $\mJ$ and
accuracy (that can be related to Peak-Signal-to-Noise-Ratio) and the
CatSIM metrics with $\kappa$, $\mAR$, $\mJ$ and accuracy. 

\begin{figure}[ht]
  \centering
  \includegraphics[width=\columnwidth]{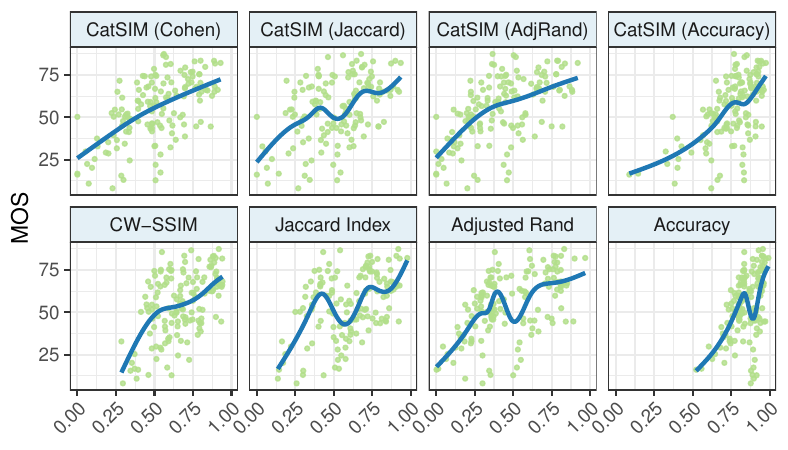}
  \caption{Adjusted Rand index, CW-SSIM, CatSIM, Jaccard index, and accuracy compared to
    mean opinion scores (MOS) rating for binary images in the first survey. CatSIM is computed using
     $\kappa$, $\mAR$, $\mJ$ and accuracy.}\label{fig:bincombined}
\end{figure}

Figure~\ref{fig:bincombined} displays the MOS with %against 
each of the metrics under consideration. These metrics are all
positively correlated with the MOS, with CatSIM methods using
$\kappa$,  $\mAR$ and accuracy performing the best (Table~\ref{tab:corrandp}\subref{tab:corr}). Given that the two categories are fairly symmetric in occurrence in the binary images, it is perhaps unsurprising that $\mJ$ and CatSIM($\mJ$) values are a bit lower, but even here the latter performs better than the  former that is the space-unaware version.  A randomization
test~(see Section~\ref{app:corrtest} for details) indicated
significantly higher correlations with the MOS for CatSIM 
($\kappa$) against CW-SSIM and accuracy and for both CatSIM ($\mAR$)
and CatSIM (Accuracy) against
accuracy~(Table~\ref{tab:corrandp}\subref{tab:pvalue}). 
%  and described in Appendix~\ref{app:corrtest}.
% While all eight metrics show positive
% association~(Table~\ref{tab:corrandp}\subref{tab:corr}) with the
% MOS, $\mJ$ or % this bit seems like we can get rid of most of it CatSIM($\mJ$) are a poor fit for a subset of the images. Here again, the goodness-of-fit measures of the CatSIM metrics are the best. 
  \begin{table}
%   \addtolength{\tabcolsep}{-3pt}
    %\begin{minipage}{0.4\columnwidth}
    \centering
      \caption{(a) Correlation with MOS for each metric and (b)
      $p$-value of test statistic for the assessment on binary images.
          \label{tab:corrandp}}
   \mbox{
     \subfloat[Correlations ($\rho$)]{
     \begin{tabular}[ht]{lc}
      \toprule
      Method & $\rho$ \\ 
      \midrule
      CatSIM (Acc) & 0.601 \\ 
       CatSIM ($\kappa$) & 0.598 \\ 
      CatSIM ($\mAR$) & 0.580 \\ 
          MS-SSIM & 0.578 \\ 
      Cohen's $\kappa$ & 0.577 \\ 
       $\mAR$ & 0.557 \\ 
       CW-SSIM & 0.510 \\ 
       Accuracy & 0.500 \\ 
      CatSIM ($\mJ$) & 0.470 \\
      Jaccard & 0.464 \\ 
      \bottomrule
     \end{tabular}
   \label{tab:corr}
   }
   \subfloat[$p$-value of test statistic that MOS correlates with CatSIM metrics (column) more than with other methods (row). Bold text indicates significance at the 5\% level.]{\label{tab:pvalue}
%     \caption{P-values for the randomization test of whether    the CatSIM methods are more correlated with MOS.}\label{tab:pvalue}
        \raisebox{0.26in}{
        \begin{tabular}{lccc}
                   \toprule
\parbox[t]{.5cm}{} & \parbox[t]{.7cm}{CatSIM ($\kappa$)} & \parbox[t]{.7cm}{CatSIM ($\mAR$)} & \parbox[t]{.7cm}{CatSIM (Acc)} \\ 
  \midrule
\!MS-SSIM & 0.311 & 0.424 & 0.410 \\ 
\!  CW-SSIM & \textbf{0.018} & 0.082 & 0.069 \\ 
\!  Accuracy & \textbf{0.032} & \textbf{0.029} & \textbf{0.033} \\ 
\!  Cohen's $\kappa$ & 0.203 & 0.369 & 0.348 \\ 
  $\mAR$ & 0.123 & 0.246 & 0.204 \\ 
   \bottomrule
 \end{tabular}
}}}

      %  \begin{figure}[H]
      %    \input{figure/corr-plot.tex}
%              \includegraphics[width=\linewidth]{figure/corr-plot-crop.pdf}
      %      \end{figure}
% \caption{Correlations of the MOS with the metrics.
%             Bold indicates a difference with CatSIM ($\kappa$), ($\mAR$) and (Acc). Italics indicate
%             a difference with  CatSIM ($\kappa$). P-values for the comparisons
%           are reported in the figure on the right.}\label{fig:corrpvalue}
    \end{table}
To assess relationships beyond linear association, we also 
fit (Figure~\ref{fig:bincombined}) a monotonic generalized additive model (GAM) to the MOS-values against each 
metric~\citep{Pya2014,scam2019} using generalized cross-validation (GCV)
to choose the GAM smoothing parameters. While the CW-SSIM model
(Table~\ref{tab:deviance}) had the 
best fit (explaining 39.1\% of the deviance), CatSIM ($\mAR$) and
CatSIM ($\kappa$) were ahead of the rest with 37.5\% and 37.1\% of the  explained deviation. We refer to Section~\ref{app:gam} for more details. 
%The full results of this analysis are available in Appendix~\ref{app:gam}. 
An interesting observation about Figure~\ref{fig:bincombined}
is that our CatSIM metrics use the entire range in [0,1], while some of the other metrics (CW-SSIM, $\mJ$ and Accuracy) seem compressed over a smaller range, therefore losing out on the ability to provide greater discrimination.
%%% Local Variables:
%%% mode: latex
%%% TeX-master: "catsim-paper"
%%% End:

\subsubsection{Assessment on Multinary Images}
\label{sec:mult}

\begin{figure}[h]
  \centering
  \includegraphics[width=\columnwidth]{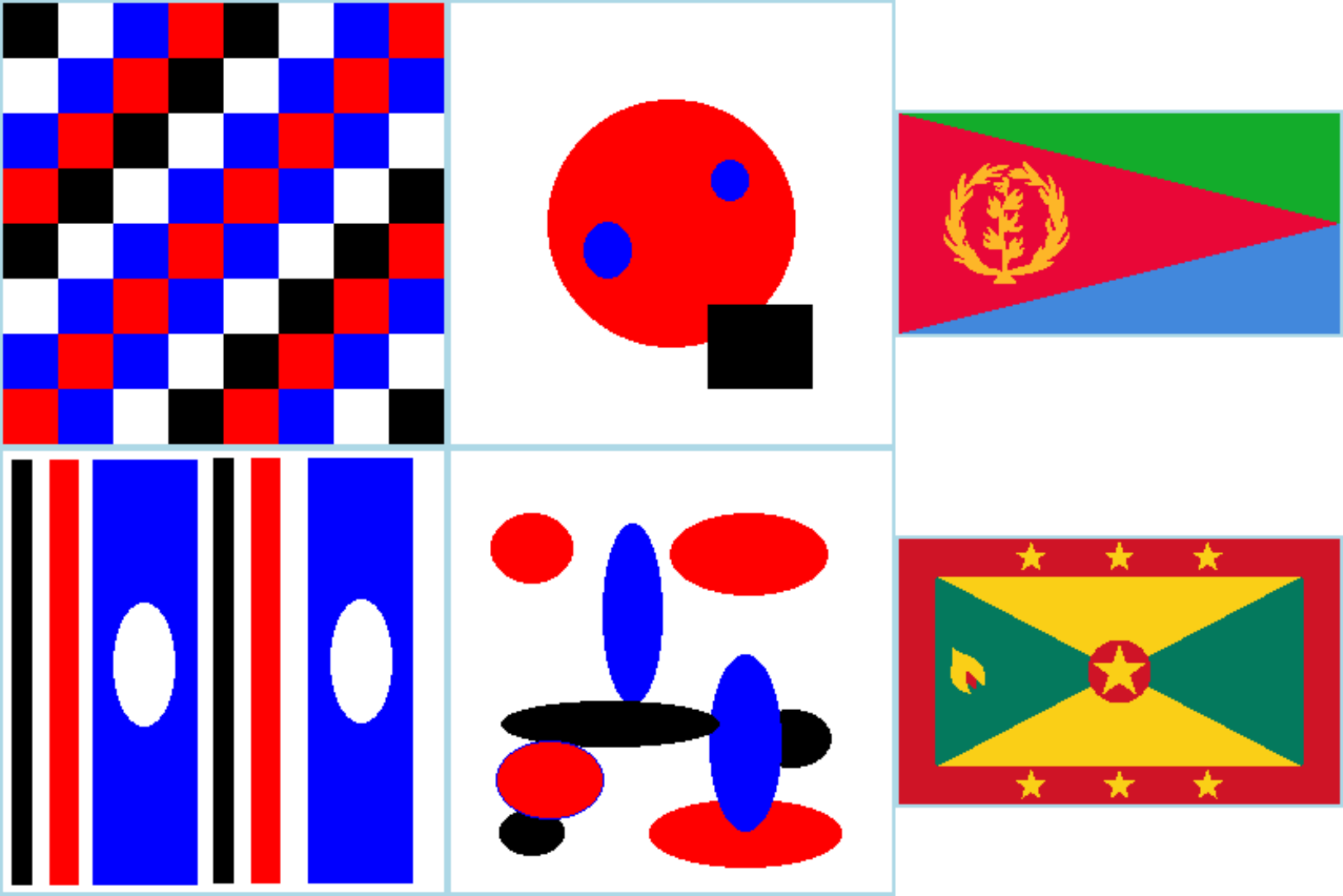}
  \caption{The undistorted multinary images used in the image ranking survey.
    Respondents were presented a series of panels of four distorted versions of each
  image and asked to rank their quality.}\label{fig:colorex}
\end{figure}

% to be inserted once ready
This survey used the six multicategory images of
Figure~\ref{fig:colorex}. 
% montage color-ex-1.png color-ex-2.png color-ex-4.png color-ex-3.png color-ex-6.png color-ex-5.png -tile x2 -geometry +0+0 -borderwidth 2 -bordercolor gray color-base-montage.pdf
Each of 614 adult volunteers were shown 11 sets of four distorted images
(see Section~\ref{app:rank} for examples)
along with the ground truth image, and were asked to rank the distorted images
from most to least similar to the original. (The sample size for this survey was larger because multinary settings require larger sample sizes. We also enhanced power by asking 
participants to  rank within each panel of four images rather than score each individual image. Ranking each image  provides a more
objective and robust basis for comparison within and across subjects, and is in any case, the basis for the Friedman~\citep{friedman37} and other similar nonparametric tests~\citep{conover80}.) 
In these examples, the labels in the original and distorted images had the same meaning, so
using CatSIM with (say) $\kappa$ rather than $\mAR$ or $\mR$ is more
appropriate. Table~\ref{tab:catranks} reports the squared difference
between the mean rankings of the sets 
of images by the raters and by the different metrics (the CatSIM
variants and the space-unaware accuracy, $\mR$, $\mAR$ and $\kappa$).  
\begin{table}
  \caption{Squared differences in mean rankings produced by human raters
    in the survey and the rankings produced by similarity metrics. }\label{tab:catranks}
\centering
\begin{tabular}[H]{lcc}
  \toprule
  Method & Squared Difference & RMSE\\  \midrule
  CatSIM (Accuracy) & 72.385 & 0.1933 \\
  CatSIM (Rand) & 74.434 & 0.1961 \\
CatSIM ($\kappa$)& 77.695 & 0.2003 \\
  MOS & 79.848 & 0.2030\\
  Adjusted Rand Index & 81.525 & 0.2052 \\
Cohen's $\kappa$ & 81.548 & 0.2052 \\
  Rand Index & 85.014 & 0.2096 \\
    Accuracy & 85.461 & 0.2101 \\
  CatSIM (Adj Rand) & 87.838 & 0.2130 \\
  \bottomrule
\end{tabular}
\end{table}
In this experiment, the CatSIM methods using
accuracy, the Rand index, or Cohen's $\kappa$ produce rankings
more similar to those produced by human raters compared to 
methods that consider only pointwise measures of agreement.

%%% Local Variables:
%%% mode: latex
%%% TeX-master: "catsim-paper"
%%% End:

\subsection{Application to Real-Data Examples}
\label{sec:application}
Our final evaluations are with two real-life applications, the
first of which computes similarity between binary 3D image volumes, while the other application was with 2D multinary images. Both our applications involve the use
of masks, which our metric effortlessly includes in its computations. 
\subsubsection{Assessing Test-Retest Reliability of Activation in  fMRI}\label{sec:fmri}
Repeatability of results across multiple fMRI~\citep{lazar08,lindquist08} studies is important to understand the variability of activation~\citep{mcnameeandlazar04,gullapallietal05} and to gauge its potential in single-subject studies~\citep{maitra09} with a view to its adoption in clinical settings. Reliability of such activation is
challenged by many factors~\citep{maitraetal02,maitra09}, not least of which is
the fact that very few (no more than 2-3\% of in-brain voxels) are expected to be truly
activated~\citep{chenandmaitra23}. \citet{maitra10} introduced $\mJ$ (instead of $\mD$) in fMRI
to more finely assess reliability of activation, and a summarized
version ($\ddot{\mJ}$) across $K$ replicated studies that 
computes the largest eigenvalue ($\lambda_{\mJ}$) of the matrix of
pairwise $\mJ$s and sets  $\ddot{\mJ} =
(\lambda_{\mJ} - 1)/(K-1)$. The underlying $\mJ$ is 
space-unaware so here we assess whether CatSIM($\mJ$) can further 
improve consistency of detected activation across multiple studies.

\begin{figure}[h]
  % \centering
  \mbox{
  \subfloat[right- (top row) and left-hand (bottom) experiments.]{\includegraphics[width=0.99\columnwidth]{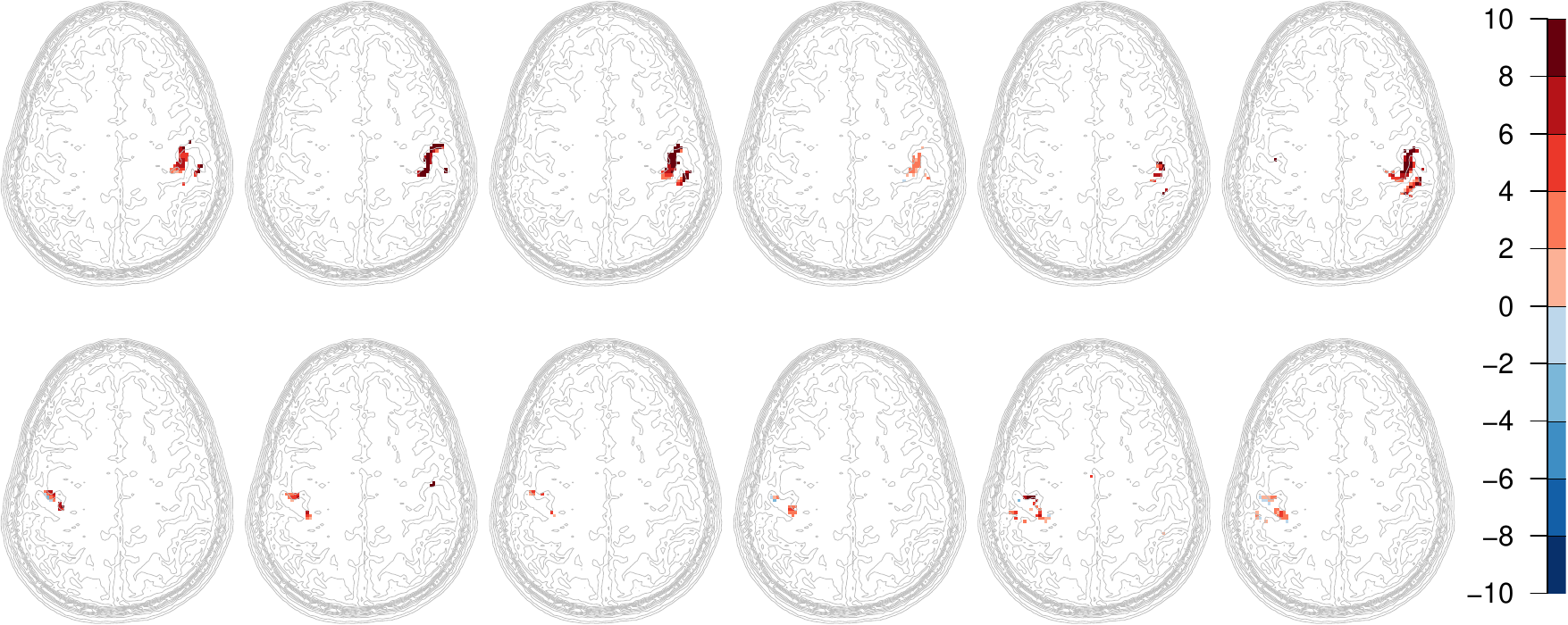}\label{fig:fmri.fast}}
}%
\\%
\mbox{
  \subfloat[$\mJ$]{\includegraphics[width=0.455\columnwidth]{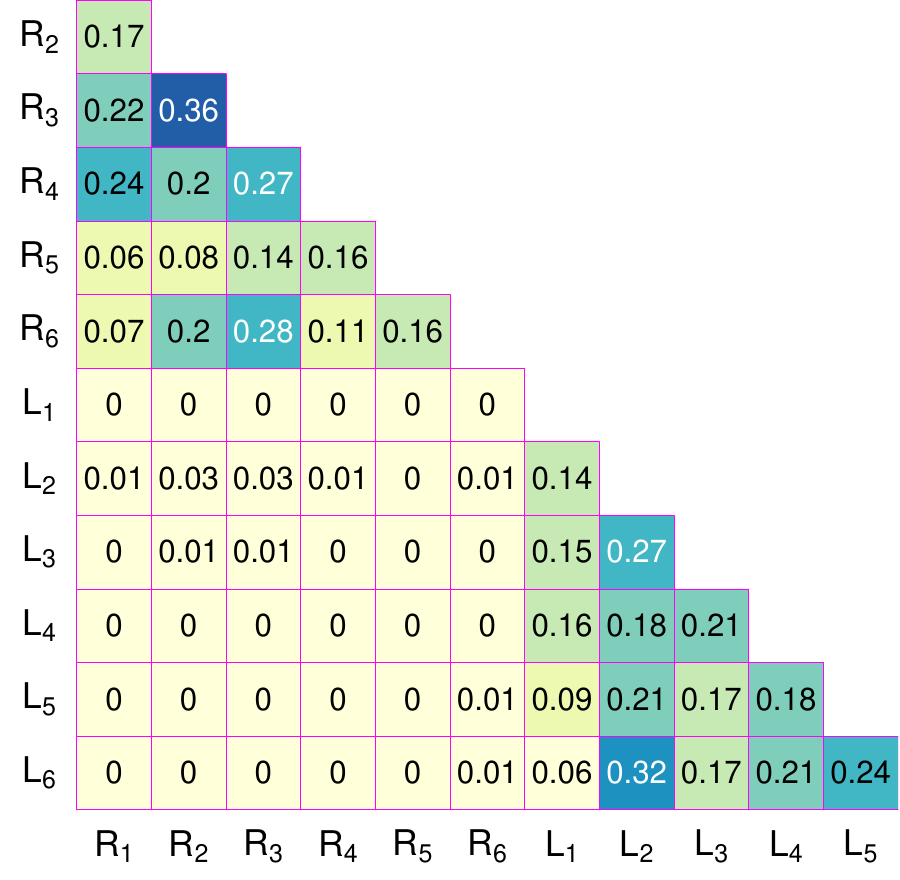}\label{fig:fmri.J}}
  \subfloat[CatSIM($\mJ$)]{\includegraphics[width=0.455\columnwidth]{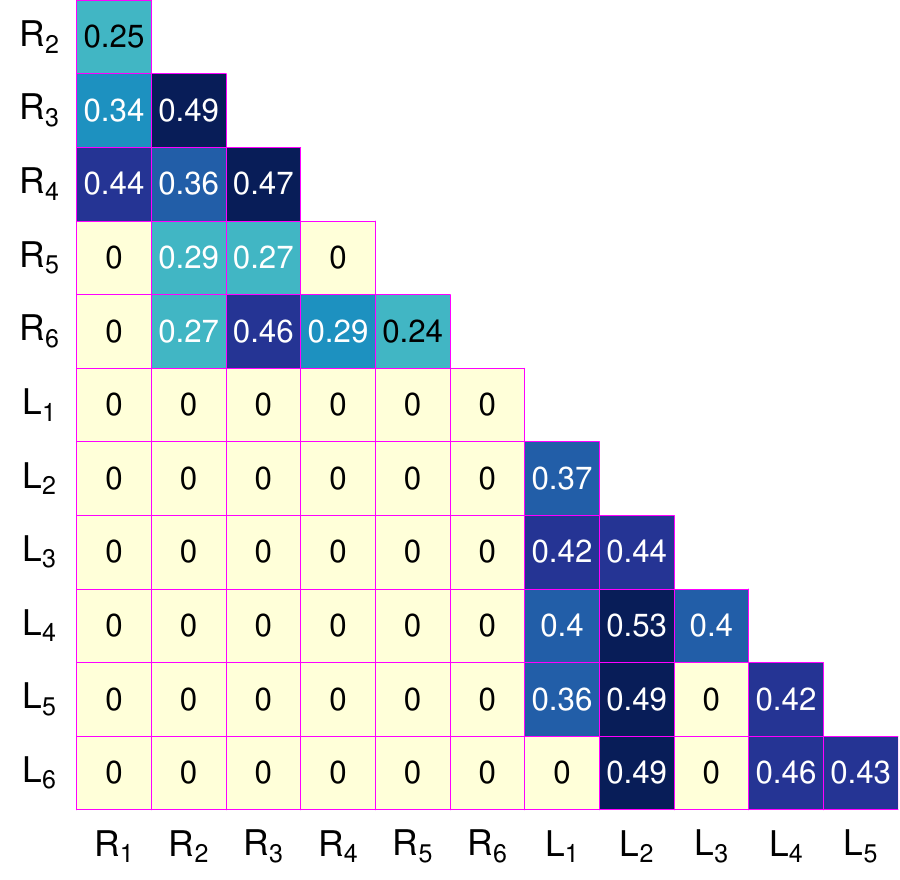}\label{fig:fmri.catsim}}
  \subfloat{\includegraphics[width=0.075\columnwidth]{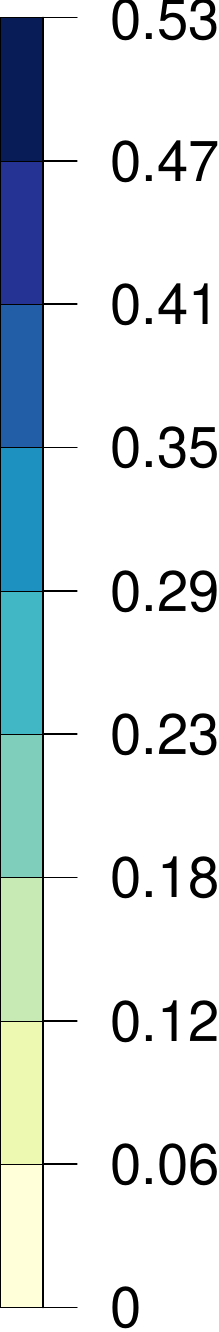}\label{fig:legend}}
}
\caption{Activation images of the 20th slice in the finger-tapping
  experiments. (b,c) Graphical displays of $\mJ$ and CatSIM($\mJ$)
  values for each 3D volume pair, with  $\mbox{R}_i$ or $\mbox{L}_i$
  indicating $i$th right- or left-hand experiment.} 
\label{fig:fmriex}
\end{figure}

Our data are from the replicated right- and left-hand finger-tapping
experiments of~\citet{maitraetal02} in which activation was detected
using the AR-FAST~\citep{almodovarandmaitra19} algorithm. For clarity of
presentation, we restrict attention only to the six
$128{\times}128{\times}22$ images that had the most
detected activation for each hand. Most of the activation (Figure~\ref{fig:fullfmri}) is,
expectedly, in the 18th through the 21st slices encompassing the (left
or right, converse to the hand used for tapping) ipsi- and
contra-lateral pre-motor cortices (pre-M1), the primary motor 
cortex (M1), the pre-supplementary motor cortex (pre-SMA) and the
supplementary motor cortex (SMA). There is wide variability of
detected activation in the contra-lateral pre-M1, pre-SMA and SMA
voxels. Figure~\ref{fig:fmriex}\subref{fig:fmri.fast} displays radiologic views of  
the activation in the 20th slice across the 12
experiments. Given the very small proportion of expected activated
voxels, we only use $\mJ$ and
CatSIM($\mJ$). Figure~\ref{fig:fmriex}\subref{fig:fmri.J} displays the
pairwise $\mJ$ between the 12 3D activation image volumes. We see
highest similarity between the second and the third activation
images. Mild similarity between detected activation in some of the
left-hand and right-hand activation maps is  also
reported. On the other hand,
Figure~\ref{fig:fmriex}\subref{fig:fmri.catsim} shows that such
commonality of activation is likely from stray voxels and not
structurally supported, as also seen by careful inspection of the
activation images in Figure~\ref{fig:fullfmri}. Further,
Figure~\ref{fig:fmriex}\subref{fig:fmri.catsim} shows that there is
greater reliability in the activation detected in the left-hand
experiments which show greater specificity as the right-hand-dominant
male shows greater focus in carrying out a left-hand task. In general,
Figure~\ref{fig:fmriex}\subref{fig:fmri.catsim} shows a wider range of
values than Figure~\ref{fig:fmriex}\subref{fig:fmri.J}, allowing for
greater discrimination. Further, $\ddot{\mJ}$, calculated using the pairwise
$\mJ$s for each of the right- and left-hand experiments were both
0.189 but 0.297 and 0.361 when the pairwise
CatSIM($\mJ$) values were substituted for $\mJ$ in the calculation of
the summarized coefficient. While the higher values for the left-hand
experiments are as expected for a right-hand-dominant subject, the
generally low values of CatSIM($\mJ$) (or $\mJ$) illustrate the
challenge of  reliable activation detection in single-subject fMRI.
\subsubsection{Evaluating Image Segmentation Algorithms}
\label{sec:brain}
\begin{figure}[h]
  \centering
  \mbox{
    \subfloat[Baseline image C080 from the BrainWeb data set.][True image]{
      \includegraphics[width=.1125\textwidth]{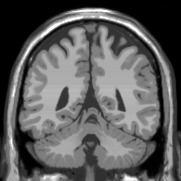}
      \label{subfig:brain1}}%
    \subfloat[Image C080 with 5\% Rayleigh noise added from the BrainWeb data set.][+5\% Noise]{
    \includegraphics[width=.1125\textwidth]{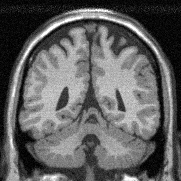}
    \label{subfig:brain2}}%
  \subfloat[Ground truth segmentation for image C080 from the BrainWeb data set.][True Labels]{
    \includegraphics[width=.1125\textwidth]{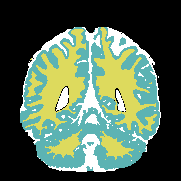}
    \label{subfig:brain3}}%
  \subfloat[Predicted SLIC-BiLSTM-MA segmentation for image C080 from the BrainWeb data set.][Prediction]{
    \includegraphics[width=.1125\textwidth]{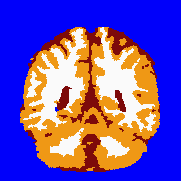}
    \label{subfig:brain4}}}%
\\%
  \mbox{
  \subfloat[]{\includegraphics[width=.5\textwidth]{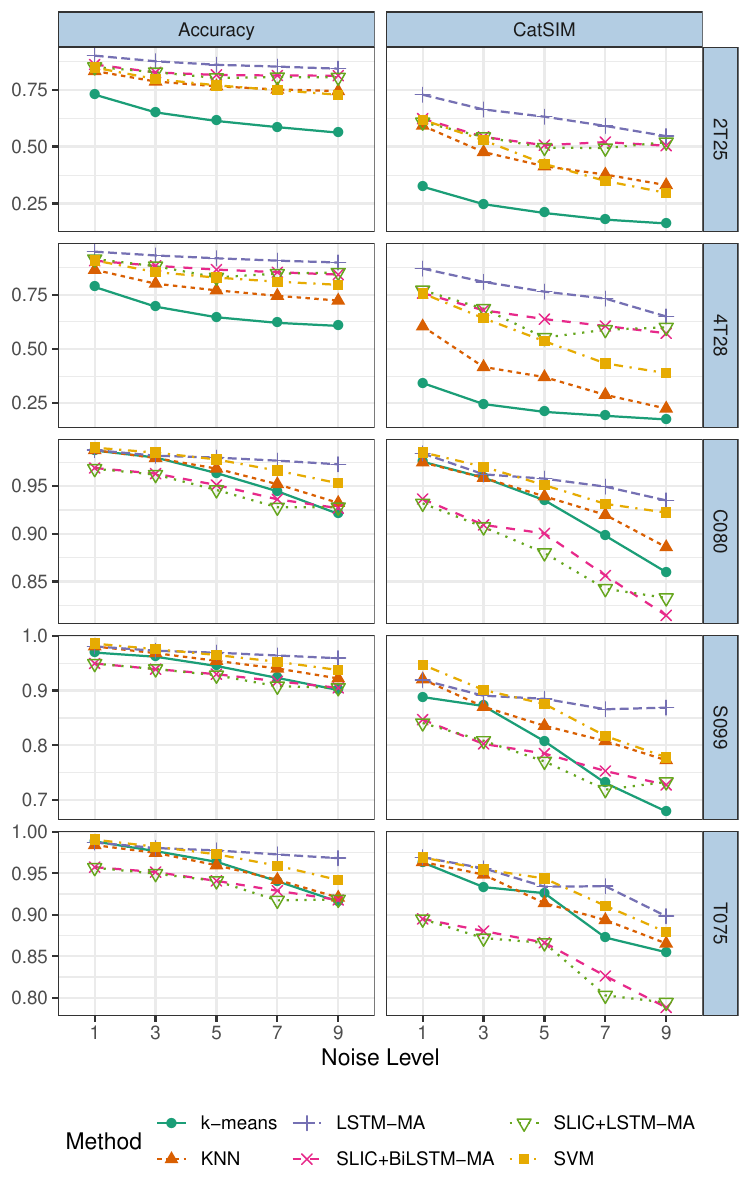}\label{fig:brainresults}}
}%
\caption{(a) Baseline image of slice C080 from the   simulated
  BrainWeb   data set and the same image with  (b) 5\% Rayleigh noise
  added, (c) ground truth segmentation and (d) predicted segmentation
  of     the noisy image using the SLIC-BiLSTM-MA algorithm. (e)
  Comparison between  Accuracy and CatSIM(Accuracy) for MR images in
  evaluating different    segmentation algorithms of MR images with
   different noise levels.}\label{fig:brainex}  
\end{figure}

Image segmentation is important in several applications with 
many algorithms whose performance needs to be calibrated. For example,
segmenting Magnetic Resonance~(MR) images into regions of gray matter, white matter, or cerebrospinal fluid is important
for diagnostic purposes and important for automated image
processing. We demonstrate CatSIM on a practical example using data
made available by \citet{XieWen19} who evaluate their new segmentation
algorithm using simulated datasets from
BrainWeb~\citep{KwanEvansPike1999,KwanEvansPike1996,Collins1998}, and
real-data images from MRBrainS~\citep{Mendrik2015}. 
%In order to evaluate their new segmentation algorithms, Xie and
%Wen~\cite{XieWen19} consider a simulated MR brain data set,
%BrainWeb~\cite{KwanEvansPike1999,KwanEvansPike1996,Collins1998}, and
%a set of brain images segmented by experts,
%MRBrainS~\cite{Mendrik2015}.
The BrainWeb interface provides multisequence ($\mbox{T}_1$-,
$\mbox{T}_2$- and  proton density-weighted) simulated brain images
with different levels of nonhomogeneous Rayleigh noise.  
The MRBrainS dataset consists of multisequence ($\mbox{T}_1$-weighted, $\mbox{T}_1$-weighted inversion recovery, and
$\mbox{T}_2$-weighted fluid attenuated inversion recovery) 3T MR scans of twenty subjects, manually-segmented
by experts. \cite{XieWen19} compare existing segmentation algorithms
to their method which uses LSTM (long short-term memory) recurrent
neural networks that account for the multi-modal nature of the data
and local structure in their classification. Here we present
comparisons of three of their methods -- LSTM-MA (LSTM with multi-modality and adjacency constraint),
SLIC-LSTM-MA (LSTM with multi-modality and super-pixel adjacency constraints),
and SLIC-BiLSTM-MA (bi-directional LSTM with multi-modality and
super-pixel adjacency constraints) -- 
to $k$-means, %fuzzy $k$-means (FCM),
support vector machines (SVM), and $k$-nearest neighbors (KNN) for
both the MRBrainS and BrainWeb data with different 
amounts of Gaussian (for MRBrainS) and Rayleigh (for Brainweb) noise. 
Figure~\ref{fig:brainex}\subref{subfig:brain1} shows an example
baseline image, Figure~\ref{fig:brainex}\subref{subfig:brain2} shows one of
its noisy versions, Figure~\ref{fig:brainex}\subref{subfig:brain3} displays 
the ground truth segmentation, and 
Figure~\ref{fig:brainex}\subref{subfig:brain4} illustrates a
segmentation of the noisy image obtained using SLIC-BiLSTM-MA.
\cite{XieWen19}  provide five examples, three (Images C080, S099, and
T075) of which are coronal, sagittal, and transverse slices from the
BrainWeb data set, and two (Images 2T25 and 4T28) are transverse
slices from the MRBrainS data set. They evaluate all  methods using
accuracy and $\mD$.   
%Images 2T25 and 4T28 are transverse slices from the MRBrainS data set
%and C080, S099, and T075 are coronal, sagittal, and transverse slices
%from the BrainWeb data sets.
Figure~\ref{fig:brainex}\subref{fig:brainresults} presents a comparison between the accuracy 
and the CatSIM metric (with accuracy) for these methods across the
different levels of added noise.
\begin{comment}
Expert segmentations 
of real brain images provide a ground truth which can be used to train image processing
models and against which image processing models can be compared. Simulated brain
data sets, where the ground truth segmentation is known, can also help evaluate
the performance of  image segmentation algorithms by comparing the results
of the classification on the simulated MR brain image with noise
added to the known ground truth. Different algorithms will produce
different segmentations which can then be evaluated against the
ground truth.
\end{comment}
\begin{comment}
\begin{figure}
  \centering
  \includegraphics[width=\columnwidth]{figure/brain-results.pdf}
  \caption{Comparison between Accuracy and CatSIM for MR brain images for
  a set of segmentation algorithms with added noise.}\label{fig:brainresults}
\end{figure}
\end{comment}
We see that the CatSIM metric gives, for almost all noise settings
and across all images, the same ordering of the methods as the accuracy. 
However, the spread of the results is greater, meaning that, as in
% the fMRI application of
Section~\ref{sec:fmri}, we get better discrimination between the
methods using CatSIM than with metrics that only assess pointwise accuracy.

\section{Discussion}
We have presented  a novel image similarity metric called CatSIM that
accounts for structural similarities in binary and multinary images and 
extends to 3D volumes. The metric can be used with masks, which means
that we can accommodate arbitrary shapes inside the images and volumes. A R~\citep{rproject} package, {\tt catsim} implements the metric. 
% binary data and on a set of multinary data
As indicated by two image quality assessment surveys, CatSIM is more similar to human perception in ranking images and provides greater discrimination between 
 segmentations than currently-used metrics. These findings are supported
by results of experiments with  simulated and real data sets
as well as from the analysis of survey data of subjective image quality ratings to guage its correspondence with human perception. 
CatSIM can also flexibly deal with labels that have meaning or
labels that are arbitrarily assigned. 
%An R~\cite{rproject} package, \texttt{catsim}, implements the algorithm.

There are a number of issues that can benefit from further
attention. For instance, we could investigate the use of smoother
 windowing  functions. Further evaluation of the weighting at
different levels of the index would also be worth pursuing. Other
refinements could include  incorporating fuzzy class labels, different
misclassification costs, or hierarchical class information
within the framework. 
% this would require reworking some of the metrics to allow weighting of
% individual points.

\bibliography{catsim}
\bibliographystyle{mytran}
\appendices
\section{Additional results for Section~\ref{subsec:illustrate}}\label{app:illus}

\subsection*{EBI Example}\label{app:besag}
Table~\ref{tab:addbinill} provides more detailed results of the demonstration
in Section~\ref{subsec:illustrationbin} of how different metrics respond to geometric distortion
and the addition of salt-and-pepper noise. In addition to the metrics reported there,
we include the results for CatSIM using accuracy and the Rand index as the similarity
metric inside it.

% latex table generated in R 3.6.2 by xtable 1.8-4 package
% Tue Mar 24 16:50:42 2020
\begin{table}[ht]
 \centering
  \caption{CatSIM and other metrics for the distorted binary images of
    the EBI.}\label{tab:addbinill}
     \addtolength{\tabcolsep}{-3pt}
\begin{tabular}{lcccccc}
 \toprule
  Metrics &  \parbox[t]{.75cm}{Horiz. Shift}  & \parbox[t]{.75cm}{S \& P Match} & \parbox[t]{.75cm}{Vert. Shift} & \parbox[t]{.75cm}{S \& P Match} &   \parbox[t]{1.25cm}{Hor. and Vert. Shift} & \parbox[t]{.75cm}{S \& P Match} \\
  \midrule
  CatSIM 5 levels    & 0.594 & 0.515 & 0.569 & 0.516 & 0.658 & 0.557  \\
   CatSIM 1 level    & 0.464 & 0.092& 0.449 & 0.090 & 0.561 & 0.110  \\ 
   CatSIM (whole)    & 0.763 & 0.769 & 0.751 & 0.756 & 0.827 & 0.832  \\ 
   CatSIM (accuracy) & 0.806 & 0.750& 0.791 & 0.752 & 0.842 & 0.777  \\ 
   CatSIM (Jaccard)  & 0.590 & 0.581& 0.571 & 0.584 & 0.647 & 0.618  \\ 
  CatSIM ($\mAR$)   & 0.479 & 0.480 & 0.440 & 0.483 & 0.538 & 0.533  \\ 
  CatSIM (Rand)      & 0.734 & 0.723& 0.714 & 0.727 & 0.771 & 0.758  \\
  MS-SSIM            & 0.670 & 0.135& 0.659 & 0.130 & 0.700 & 0.170 \\
  CW-SSIM            & 0.831 & 0.783& 0.752 & 0.780 & 0.834 & 0.810 \\
  Accuracy           & 0.898 & 0.898& 0.893 & 0.893 & 0.926 & 0.927  \\ 
  Jaccard            & 0.720 & 0.734& 0.708 & 0.723 & 0.788 & 0.799  \\ 
  AdjRand            & 0.627 & 0.630& 0.610 & 0.613 & 0.720 & 0.725  \\ 
  Rand               & 0.817 & 0.818& 0.809 & 0.809 & 0.863 & 0.864  \\ 
  Cohen's $\kappa$   & 0.763 & 0.771& 0.751 & 0.759 & 0.827 & 0.834  \\ 
   \bottomrule
\end{tabular}
\end{table}

Table~\ref{tab:addcatill} provides more detailed results of the demonstration
in Section~\ref{subsec:illustrationmulti}. In addition to the metrics reported there,
we include the results for CatSIM using accuracy and the Rand index as the similarity
metric inside it. 
% latex table generated in R 3.6.2 by xtable 1.8-4 package
% Tue Mar 24 15:49:26 2020
\begin{table}[ht]
  \centering
  \caption{CatSIM and other metrics for the 4-class image example.}\label{tab:addcatill}
     \addtolength{\tabcolsep}{-3pt}
\begin{tabular}{lcccccc}
 \toprule
  Metrics &  \parbox[t]{.75cm}{Horiz. Shift}  & \parbox[t]{.75cm}{S \& P Match} & \parbox[t]{.75cm}{Vert. Shift} & \parbox[t]{.75cm}{S \& P Match} &   \parbox[t]{1.25cm}{Hor. and Vert. Shift} & \parbox[t]{.75cm}{S \& P Match} \\
  \midrule
  CatSIM  5 levels  & 0.816 & 0.610 & 0.652 & 0.548 & 0.814 & 0.565  \\ 
  CatSIM 1 level    & 0.686 & 0.105 & 0.462 & 0.081 & 0.533 & 0.092  \\ 
  CatSIM (whole)    & 0.906 & 0.906 & 0.827 & 0.829 & 0.869 & 0.869  \\ 
  CatSIM (accuracy) & 0.892 & 0.808 & 0.796 & 0.764 & 0.906 & 0.778  \\
  CatSIM ($\mAR$)  & 0.783 & 0.604 & 0.592 & 0.538 & 0.785 & 0.558  \\ 
  CatSIM (Rand)     & 0.879 & 0.799 & 0.770 & 0.748 & 0.886 & 0.765  \\ 
  Accuracy          & 0.936 & 0.935 & 0.881 & 0.882 & 0.910 & 0.910  \\ 
  AdjRand           & 0.828 & 0.842 & 0.723 & 0.723 & 0.777 & 0.784  \\ 
  Rand              & 0.926 & 0.933 & 0.881 & 0.883 & 0.904 & 0.909  \\ 
  Cohen's $\kappa$  & 0.906 & 0.907 & 0.827 & 0.831 & 0.869 & 0.871  \\ 
   \bottomrule
\end{tabular}
\end{table}
\section{Additional Results and Details for Section~\ref{sec:bin}}
\subsection{Randomization Test for Difference in Correlations}\label{app:corrtest}

To test whether the correlation of a metric, $m_1$, with the MOS (mean opinion score on the survey),
$M$, is greater
than another metric, $m_2$, we standardize $m_1$, $m_2$, and $M$ and perform the
following randomization test:
\begin{enumerate}
\item For each distorted image, swap standardized elements of $m_1$ and $m_2$
  with probability $0.5$ to create new vectors $\tilde{m}_1$ and $\tilde{m}_2$.
\item Compute and record $corr(M, \tilde{m}_1) - corr(M, \tilde{m}_2)$.
\item Repeat $n$ times.
\item Define the $p$ value as the proportion of times
  $ corr(M, {m_1}) - corr(M, {m_2}) > corr(M, \tilde{m}_1) - corr(M, \tilde{m}_2)$.
\end{enumerate}
The results of this test with $n = 100,000$ are listed in Table~\ref{tab:corrtest}.
% latex table generated in R 3.6.3 by xtable 1.8-4 package
% Thu Apr 16 20:33:52 2020
  \begin{table}[ht]
    \caption{Results of a randomization test of whether the correlation of
      MOS with the CatSIM method is greater than the comparison methods.
      Italics indicate $p < 0.05$. The left column indicates the CatSIM variant
    and the middle column indicates the metric it is compared to.}\label{tab:corrtest}
\centering
\begin{tabular}{rllr}
  \toprule
CatSIM metric & Comparison & p-value \\ 
  \midrule
CatSIM ($\kappa$) & MS-SSIM & 0.311 \\ 
CatSIM ($\kappa$) & CW-SSIM & \emph{0.018} \\ 
CatSIM ($\kappa$) & Accuracy & \emph{0.032} \\ 
CatSIM ($\kappa$) & Cohen's $\kappa$ & 0.203 \\ 
CatSIM ($\kappa$) & AdjRand & 0.123 \\ 
CatSIM ($\mathcal{J}$) & Jaccard & 0.374 \\ 
CatSIM (Acc) & MS-SSIM & 0.424 \\ 
CatSIM (Acc) & CW-SSIM & 0.082 \\ 
CatSIM (Acc) & Accuracy & \emph{0.029} \\ 
 CatSIM (Acc) & Cohen's $\kappa$ & 0.369 \\ 
 CatSIM (Acc) & AdjRand & 0.246 \\ 
 CatSIM ($\mAR$) & MS-SSIM & 0.410 \\ 
 CatSIM ($\mAR$) & CW-SSIM & 0.069 \\ 
 CatSIM ($\mAR$) & Accuracy & \emph{0.033} \\ 
 CatSIM ($\mAR$) & Cohen's $\kappa$ & 0.348 \\ 
 CatSIM ($\mAR$) & AdjRand & 0.204 \\ 
   \bottomrule
\end{tabular}
\end{table}

\subsection{Image rankings by  monotonic GAM}\label{app:gam}
We fit monotonic GAMs with an identity link function and Normal random
components using the metrics as explanatory variables and the MOS (mean opinion score)
as the response variable. The percentage of deviance explained by the
model, reported in Table~\ref{tab:deviance}, is then a non-parametric measure of the correspondence
between the image quality metrics and the subjective quality
of the images.
% latex table generated in R 3.6.3 by xtable 1.8-4 package
% Sat Apr 18 22:34:31 2020
\begin{table}[ht]
  \centering
  \caption{The percent of deviance in MOS explained by a monotonic GAM
  with an identity link function and normal errors with smoothing parameter
  selection by generalized cross validation. 
  }\label{tab:deviance}
\begin{tabular}{lr}
  \toprule
 Method & \% Deviance \\ 
  \midrule
CatSIM (Acc) & 0.221 \\ 
Adjusted Rand & 0.323 \\ 
Cohen's $\kappa$ & 0.329 \\ 
CatSIM ($\mJ$) & 0.334 \\ 
Jaccard & 0.348 \\ 
MS-SSIM & 0.356 \\ 
  Accuracy & 0.361 \\
  CatSIM ($\kappa$) & 0.371 \\ 
CatSIM ($\mAR$) & 0.375 \\ 
CW-SSIM & 0.391 \\ 
   \bottomrule
\end{tabular}
\end{table}

\section{Additional Details for Section~\ref{sec:mult}}
\subsection*{Image rankings in the categorical survey}\label{app:rank}
Respondents were shown 11 sets of four distorted images. For each set of
four distorted images, they were shown the true, undistorted image
and asked to rank the four distorted images by their quality compared to the true image.
The 11 sets were presented in a random order. Within the panels of
four images, the order and labels (A, B, C, and D) were fixed.
Below are the 11 sets of four distorted images shown in the survey.
There were two different sets
for two of the original images and one set for one of the images.
\begin{figure}
  \centering
  \mbox{
  \subfloat[Set 1-1.]{\includegraphics[width=.47\columnwidth]{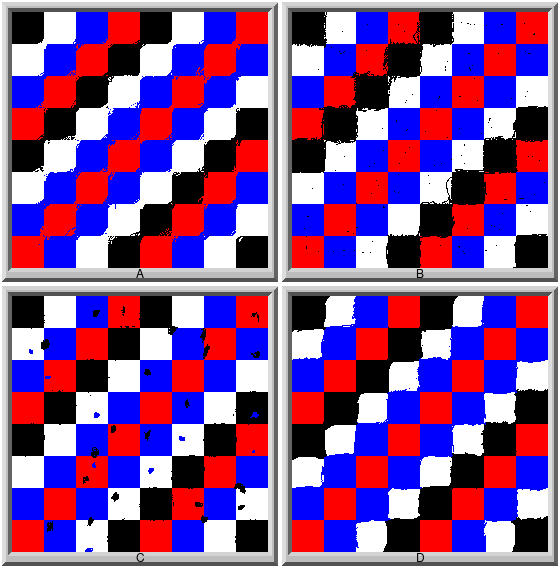}
  \label{subfig:distortoneone}}%
~%
\subfloat[Set 1-2.]{  \includegraphics[width=.47\columnwidth]{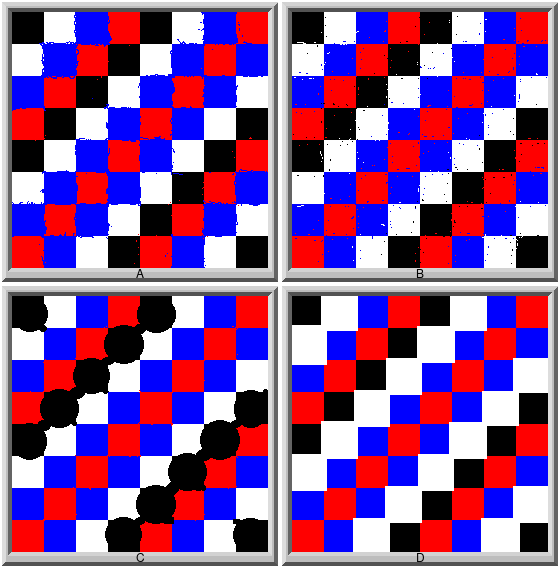}
  \label{subfig:distortonetwo}}}
\mbox{
  \subfloat[Set 2-1.]{  \includegraphics[width=.47\columnwidth]{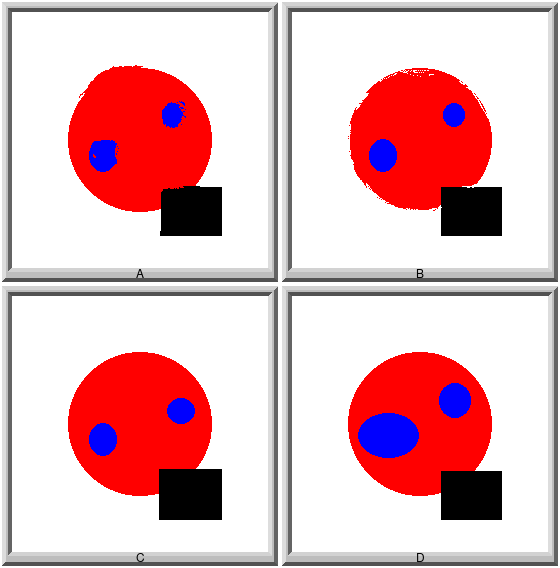}%
    \label{subfig:distorttwoone}}%
~%
\subfloat[Set 2-2.]{  \includegraphics[width=.47\columnwidth]{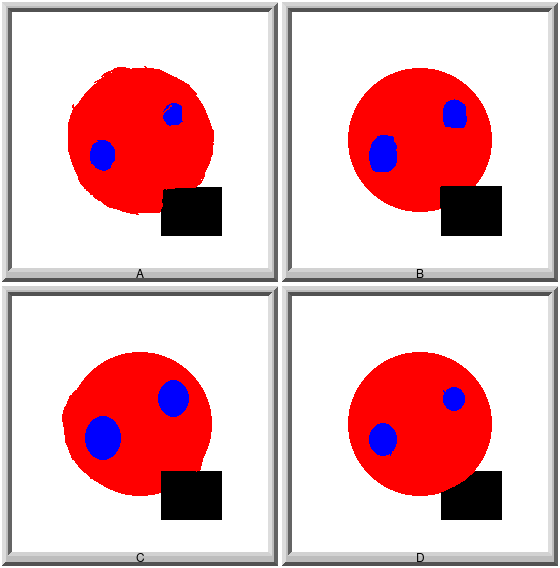}
  \label{subfig:distorttwotwo}}
}
\mbox{
  \subfloat[Set 3-1.]{  \includegraphics[width=.47\columnwidth]{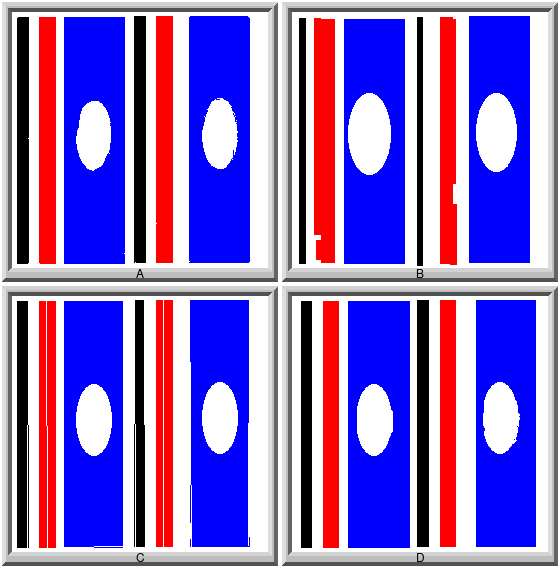}
    \label{subfig:distortthreeone}}%
~%
\subfloat[Set 3-2.]{  \includegraphics[width=.47\columnwidth]{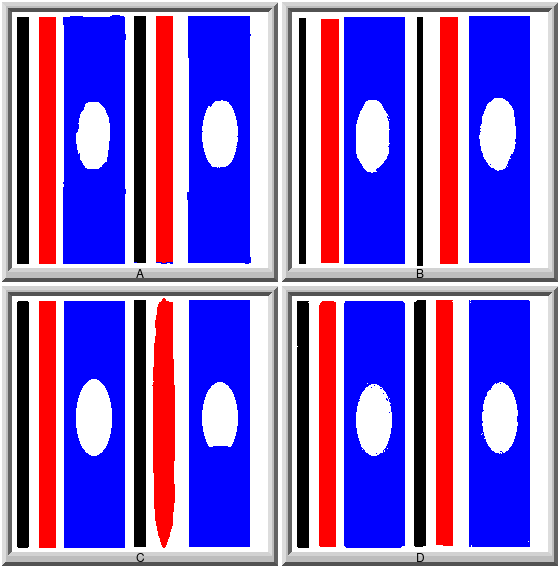}
  \label{subfig:distortthreetwo}}%
}
\mbox{
  \subfloat[Set 4-1.]{\includegraphics[width=.47\columnwidth]{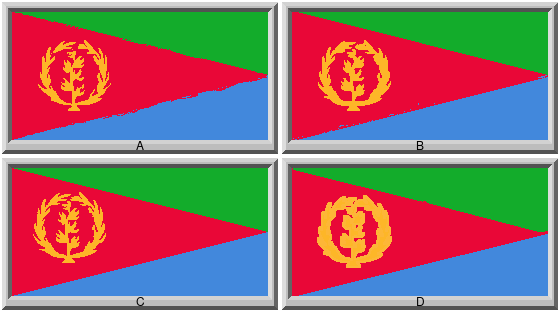}
\label{subfig:distortfourone}}%
~%
\subfloat[Set 4-2.]{\includegraphics[width=.47\columnwidth]{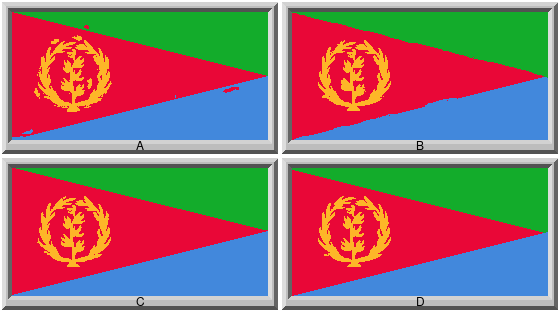}
  \label{subfig:distortfourtwo}}%
}
\mbox{
\subfloat[Set 5-1.]{\includegraphics[width=.47\columnwidth]{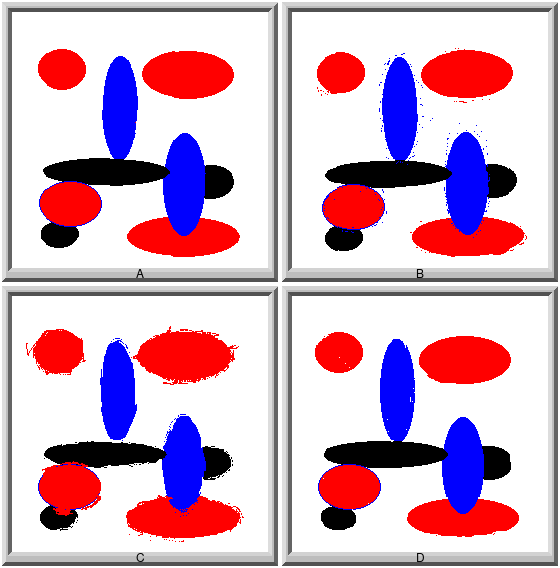}
  \label{subfig:distortsixone}}%
~%
\subfloat[Set 5-2.]{  \includegraphics[width=.47\columnwidth]{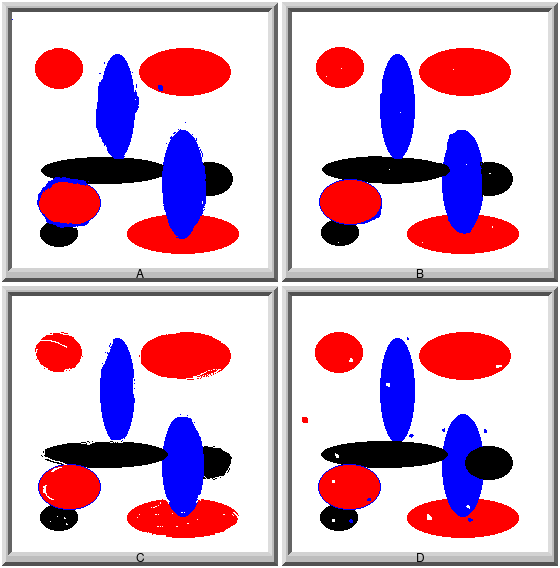}
  \label{subfig:distortsixtwo}}
}
\caption{The two sets of distorted images used for each of Images 1--5 in the image ranking survey.}\label{fig:distort}
\end{figure}

\begin{figure}
  \centering
\includegraphics[width=0.85\columnwidth]{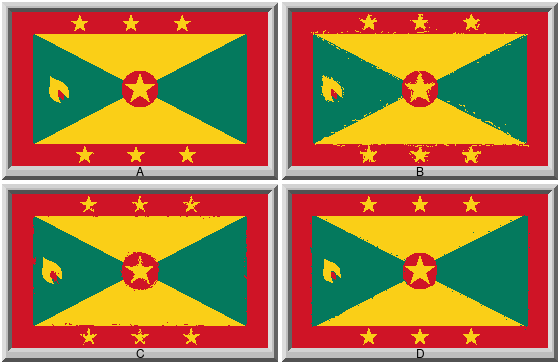}
  %\caption{Set 5-1.}
\caption{The one set of distorted images used in the image ranking survey for the sixth image.}\label{fig:distortfifth}
\end{figure}

\section{Additional details for Section~\ref{sec:fmri}}
\begin{figure}
  \vspace{-0.5in}
  \centering
\includegraphics[width=\columnwidth]{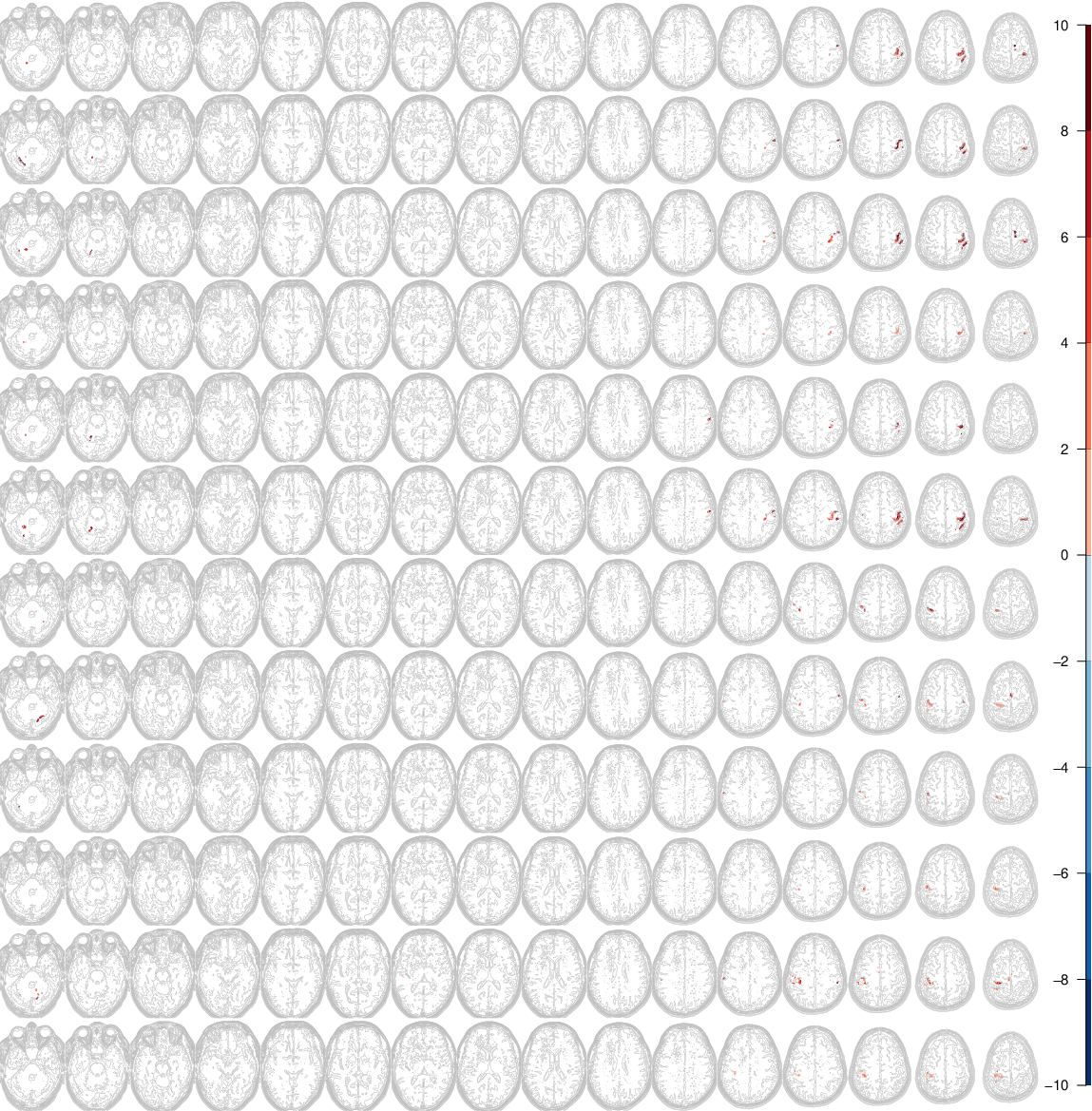}
%\caption{Set 5-1.}\label{subfig:distortfiveone}
\caption{Radiologic views of activation detected by AR-FAST in the top
  16 slices 
  (Slices 7-22, column-wise) and in each of the 6 right-hand (first six
  rows) and 
  left-hand (last six rows) finger-tapping experiment.}\label{fig:fullfmri}
\end{figure}

Figure~\ref{fig:fullfmri} provides
all the activation maps detected on the finger-tapping experiment of Section~\ref{sec:fmri} using the AR-FAST algorithm of \citet{almodovarandmaitra19}.

%%% Local Variables:
%%% mode: latex
%%% TeX-master: "supplement"
%%% End

\end{document}